\documentclass[11pt]{article}

\usepackage[margin=1in]{geometry}

\usepackage{algorithmic}
\usepackage{textcomp}
    
\usepackage{subfiles}
\usepackage{paralist}
\usepackage[mathscr]{eucal} %% For \mathscr
\usepackage{algorithmic}

\usepackage[noend,ruled,vlined,linesnumbered,resetcount]{algorithm2e}
\usepackage{booktabs}
\usepackage{graphicx}
\usepackage[T1]{fontenc}
\usepackage{textcomp}
\usepackage{comment}
\usepackage{hyperref}
\usepackage{pifont}
\usepackage{makecell}
\usepackage{balance}
\usepackage{multirow}

\usepackage{multirow}
\usepackage{tabularx,stackengine,collcell}
\let\endminwd\relax
\newcolumntype{L}[1]{>{\collectcell\xminwd l{#1}}l<{\endminwd\endcollectcell}}
\newcolumntype{C}[1]{>{\collectcell\xminwd c{#1}}c<{\endminwd\endcollectcell}}
\newcolumntype{R}[1]{>{\collectcell\xminwd r{#1}}r<{\endminwd\endcollectcell}}
\def\minwd#1#2#3\endminwd{\stackengine{0pt}{#3}{\rule{#2}{0pt}}{O}{#1}{F}{F}{L}}
\newcommand\xminwd[1]{\minwd#1}

\usepackage[center]{subfigure}
\usepackage{caption}
\usepackage{enumitem}
\usepackage{amsfonts}
\usepackage[table,dvipsnames]{xcolor}  % Add in preamble if not already
\usepackage{bbm}

\AtBeginDocument{%
  }

\usepackage{amsmath}

\newcommand{\mask}{[\mathrm M]}

\definecolor{Gred}{RGB}{219, 50, 54}
\definecolor{Ggreen}{RGB}{75, 200, 100}
\definecolor{Gblue}{RGB}{72, 133, 237}
\definecolor{Gyellow}{RGB}{247, 178, 16}
\definecolor{ToCgreen}{RGB}{0, 128, 0}
\definecolor{myGold}{RGB}{231,141,20}
\definecolor{myBlue}{rgb}{0.19,0.41,.65}
\definecolor{myPurple}{RGB}{175,0,124}

\usepackage{graphicx}

\usepackage{hyperref}

\hypersetup{
    colorlinks=true,     % required to color text instead of boxes
    citecolor=ToCgreen,  % reference color (green text)
    linkcolor=Sepia,
    filecolor=Gred,
    urlcolor=Gred
}

\newcommand{\E}{\mathbb{E}}
\newcommand{\citet}{\cite}

\title{Masked Diffusion for Generative Recommendation}
\date{}

\usepackage{adjustbox}

\author{
    Kulin Shah \thanks{Email: \texttt{kulinshah@utexas.edu}. Work done during an internship at Snap.} \\
    UT Austin 
        \and
    Bhuvesh Kumar \thanks{Email: \texttt{bkumar@snap.com}.} \\
    Snap Inc.
        \and
    Neil Shah \thanks{Email: \texttt{nshah@snap.com}.} \\
    Snap Inc.
        \and
    Liam Collins \thanks{Email: \texttt{lcollins2@snap.com}.}\\
    Snap Inc. 
}

\DeclareMathOperator{\Unif}{Unif}

\begin{document}

\maketitle

\begin{abstract}
    Generative recommendation (GR) with semantic IDs (SIDs) has emerged as a promising alternative to traditional recommendation approaches due to its performance gains, capitalization on semantic information provided through language model embeddings, and inference and storage efficiency \cite{rajput2023recommender, deng2025onerec}. Existing GR with SIDs works frame the probability of a sequence of SIDs corresponding to a user's interaction history using autoregressive modeling.
    While this has led to impressive next item prediction performances in certain settings, these autoregressive GR with SIDs models suffer from expensive inference due to sequential token-wise decoding, potentially inefficient use of training data and bias towards learning short-context relationships among tokens.
    Inspired by recent breakthroughs in NLP \cite{sahoo2024simple, shi2025simplified, ou2024absorbing}, we propose to instead model and learn the probability of a user's sequence of SIDs using masked diffusion. Masked diffusion employs discrete masking noise to facilitate learning the sequence distribution, and models the probability of masked tokens as conditionally independent given the unmasked tokens, allowing for parallel decoding of the masked tokens. We demonstrate through thorough experiments that our proposed method consistently outperforms autoregressive modeling. This performance gap is especially pronounced in data-constrained settings and in terms of coarse-grained recall, consistent with our intuitions. Moreover, our approach allows the flexibility of predicting multiple SIDs in parallel during inference while maintaining superior performance to autoregressive modeling. Our code is available at \url{https://github.com/snap-research/MaskGR}.
\end{abstract}

\bibliographystyle{alpha}
%%
%% This command processes the author and affiliation and title
%% information and builds the first part of the formatted document.
\maketitle

% \section{Introduction}

\section{Introduction}

Generative Recommendation (GR) is a rapidly growing paradigm in Recommendation Systems (RecSys) that aims to leverage generative models to recommend items to users based on users' historical interaction sequences. 
% Often, these models are encoder-decoder or decoder-only transformers trained autoregressively, or continuous diffusion models that diffuse over embedding space. 
% Despite both of these styles of models being developed outside of RecSys -- in NLP and CV -- their applications to recommendation have led to impressive results.
% Both of these types of models are motivated by their successful applications in other domains -- natural language processing (NLP) and Vision, and both have achieved impressive results in recsys...
% GR draws heavily from recent advances in natural language processing (NLP), as the most popular GR approach -- GR with semantic IDs (SIDs) -- incorporates both item text embeddings from a pre-trained language model and the autoregressive training procedure that is commonplace in NLP. 
% GR draws heavily from recent advances in natural language processing (NLP), as the most popular GR approach -- GR with semantic IDs (SIDs) -- incorporates both item text embeddings from a pre-trained language model and the autoregressive training procedure that is commonplace in NLP. 
Among GR paradigms, GR with semantic IDs (SIDs) \cite{rajput2023recommender, hou2023learning} has garnered widespread interest and usage, including successful industrial applications in long \cite{singh2024better} and short \cite{deng2025onerec, zhou2025onerec} video recommendation,  music recommendation \cite{penha2025semantic}, and online retail \cite{luo2024qarm, ma2025grace}.

The GR with SIDs framework has led to such successes in large part by offering a means to incorporate both semantic and collaborative signals in item representations that use an exponentially smaller vocabulary size than traditional sparse IDs \cite{rajput2023recommender, yang2024unifying}. These representations are learned by first executing {\em item SID assignment}: assigning tuples of tokens, known as SIDs, to items based on a residual clustering of their text and/or visual feature embeddings extracted from a pretrained language and/or vision model. Then, in the {\em user sequence modeling} phase, embeddings of these tokens are trained in concert with a sequential model to learn the probability distribution of the sequences of SIDs corresponding to users' interaction histories \cite{rajput2023recommender, grid}.
% GR with SIDs entails a tokenization procedure to assign SIDs to items, followed by an autoregressive training to learn the probability distribution of the sequences of SIDs corresponding to users' item interaction histories. 
% During tokenization, tuples of tokens, i.e. SIDs, are assigned to items based on a residual clustering of their modality feature embeddings extracted from a pretrained language or vision model. 
% Representing items with SIDs has the dual advantages of capturing semantic relationships between items and doing so with exponentially smaller codebook size than assigning a single unique ID to each item.
% After item tokenization with SIDs, a decoder-based transformer is trained to predict the next SID token in user interaction sequences in an autoregressive manner \cite{rajput2023recommender}.
Despite tremendous interest in GR with SIDs, the majority of studies focus on design choices around item SID assignment \cite{xiao2025progressive, hua2023index, qu2025tokenrec, wang2024learnable, wang2024eager, liu2024end, tan2024idgenrec, jin2023language, zhu2024cost, kuai2024breaking, houactionpiece}, adopting the default autoregressive (AR) modeling  from \citet{rajput2023recommender} for the user sequence modeling stage.

% leaving recommendation training, especially design choices around autoregressive training and inference, overlooked. 
% Surprisingly, 
% While a plethora of  works have studied tokenization design choices, 
AR modeling has long been the dominant paradigm  in Natural Language Processing (NLP) \cite{rumelhart1985feature, bengio2003neural, sutskever2014sequence}, leading to countless breakthroughs in the development of LLMs \cite{vaswani2017attention, radford2018improving, radford2019language, brown2020language}. This makes it a reasonable starting point for modeling SID sequences, and indeed, existing results verify that AR modeling's efficacy for modeling text sequences translates to modeling users' SID sequences. However, since AR models model the probability of all future tokens as dependent on all preceding tokens, they must generate tokens sequentially, leading to fundamental upper bounds on inference speed and planning. Moreover, in practice, by training on next token prediction, AR models tend to under-index on global relationships among tokens \cite{khandelwal2018sharp, henighan2020scaling, brunato2023coherent}.
% \cite{lin1996learning}

% This is problematic in light of recent results in 
% for GR with SIDs \cite{}, surprisingly little attention has been paid to the recommendation stage.
% The standard autoregressive training procedure employed by most GR with SIDs methods is borrowed from NLP, where it has been the default LLM training strategy for several years \cite{}.
% However, more recently, the dominance of autoregressive training in NLP has been called into question by an alternative method:  masked, or discrete, diffusion \cite{sahoo2024simple}. Masked diffusion has been shown to perform... \cite{} and has been adopted for... \cite{}
% {\color{red}
Diffusion modeling in NLP alleviates these issues by modeling the probability of future  tokens as independent conditioned on the context tokens, allowing for parallel token generation.
% NLP alleviates these issues  by training on a masked language modeling objective 
Early approaches to diffusion language modeling applied continuous, i.e. Gaussian, diffusion over token embeddings
\cite{li2022diffusion, gong2022diffuseq, yuan2022seqdiffuseq, zhou2023denoising}, mimicking the continuous diffusion protocol that has a long track record of success in image generation \cite{sohldickstein2015deep, ho2020denoising, dhariwal2021diffusionmodelsbeatgans, rombach2022high, ho2022classifier}, but failed to 
 match the performance of AR models on language modeling benchmarks \cite{lou2024discrete, gong2024scaling}. However, more recent advancements have shown that an alternative diffusion paradigm -- {\em masked diffusion} -- can compete with AR language modeling on standard benchmarks \cite{austin2023structured, lou2024discrete, sahoo2024simple, shi2025simplified, ou2024absorbing}. The key innovation is the use of discrete masking noise rather than Gaussian noise, which empirically leads to improved learning of discrete token sequence distributions.
% is only due to recent advancements that they have achieved general text generation performance on par with that of AR modeling . A 

% discrete diffusion 
% \cite{sahoo2024simple}
% \cite{shi2025simplified}
% \cite{ou2024absorbing}

% \cite{austin2023structured} (generalization of masked diffusion)
% \cite{lou2024discrete}
% \cite{sun2022score}

% continuous diffusion for text
% \cite{li2022diffusion}
% \cite{yuan2022seqdiffuseq}
% \cite{gong2022diffuseq}
% \cite{zhou2023denoising}

% \cite{he2022diffusionbert}
% \cite{zheng2023reparameterized}

% success of diffusion in cv
% \cite{sohldickstein2015deep, ho2020denoising}
% \cite{dhariwal2021diffusionmodelsbeatgans, rombach2022high, ho2022classifier}

% In masked diffusion, at each training step, we randomly sample a noise ratio $t$, randomly mask $t$ tokens per sequence, and train an encoder to predict the masked tokens. During inference, we mask the future tokens, and ask the model to predict those tokens. 
% Masked diffusion offers several advantages over autoregressive training.

This masked diffusion framework is especially promising for GR over discrete SID sequences for several reasons.
First, by employing random masking with all possible masking rates, masked diffusion uses exponentially many training samples per raw sequence in the raw sequence length, whereas  AR modeling uses only linearly many. 
In NLP, this more aggressive augmentation strategy tends to extract more signal out of limited training data \cite{prabhudesai2025diffusion, ni2025difflm}, which is appealing for recommendation scenarios that typically have sparse interactions between users and items.
Second, masked diffusion may better capture global relationships among tokens, by virtue of being trained to unmask throughout the sequence rather than continually predict the next token.
Third, masked diffusion allows for simultaneously decoding multiple tokens in one forward pass of the model.
% unlike AR models which must make one forward pass per generated token, 
This is critical for GR with SIDs  since decoding an item requires decoding multiple SID tokens, and inference latency is often a bottleneck for practical deployment \cite{he2014practical}.
% }

% These advantages warrant the 

% the superiority of  autoregressive training compared to alternative training methods, especially discrete diffusion, has been called into question by multiple works. ... showed on several benchmarks that discrete diffusion extracts more generalizable information per unit of training data given sufficiently many training epochs. ...
% Despite these advantages, to the best of our knowledge, a thorough study of masked diffusion for GR does not yet exist in the literature. In the is work, we address this gap by
% These observations make the lack of innovation on autoregressive training in GR with SIDs even more alarming.
% In this work, we address this gap by 
% developing and studying the first method for masked diffusion over SIDs. we call our method ...
In this work, we realize these advantages for GR with SIDs by developing \textbf{MaskGR}: \textbf{Mask}ed Diffusion over SIDs for \textbf{G}enerative \textbf{R}ecommendation. To the best of our knowledge, MaskGR is the first application of  masked diffusion for GR with SIDs, and 
% implementation of masked diffusion for GR. We develop MaskGR by... // MaskGR entails...
% Through thorough experiments, we make the following intriguing observations about MaskGR:
as such, MaskGR prioritizes simplicity and generality in its design.
This allows us to make a number of foundational empirical observations regarding the effectiveness of masked diffusion for GR with SIDs:
\begin{itemize}[leftmargin=*]
    \item \textbf{Overall performance.} MaskGR consistently outperforms standard AR modeling with SIDs (i.e. TIGER \cite{rajput2023recommender}) as well as other GR baselines on several benchmark sequential recommendation datasets. The performance improvement is especially large for coarse-grained recall, suggesting masked diffusion's strength at learning global item relationships.
    \item \textbf{Data efficiency.} Up until a point at which the data becomes too sparse, the performance gap between MaskGR and TIGER grows as we shrink the dataset size, supporting the hypothesis that masked diffusion makes better use of limited data.
    \item \textbf{Inference efficiency and flexibility.} MaskGR can decode multiple tokens in parallel, which allows for flexibly trading  off inference performance and efficiency by choosing the number of forward passes (function evaluations) to execute. 
    MaskGR already outperforms TIGER with fewer function evaluations, and its performance improves with additional evaluations.
    % uses  fewer function evaluations during inference than TIGER since it can decode multiple tokens simultaneously. We can further improve MaskGR's performance with additional function evaluations.
    \item \textbf{Extensibility.} MaskGR is compatible with auxiliary methods for improving performance in GR. Here we focus on incorporating dense retrieval, motivated by the impressive performance of fusing TIGER with  dense retrieval \cite{yang2024unifying, yang2025sparse}. 
    % We observe that this enhances performance by \%.
     
    % \item On the negative side, MaskGR requires more  training iterations than TIGER, consistent with comparisons of autoregressive training and masked diffusion in NLP. It also scales only slightly better the TIGER with additional SIDs per item. 

    % learning rate sensitivity ablation?
\end{itemize}

\noindent {\textbf{Reproducibility.} Our code is available at the following link:  \url{https://github.com/snap-research/MaskGR}}.

\section{Generative Recommendation with Semantic IDs}

In this section we formally introduce the Generative Recommendation with Semantic IDs  framework. We start by discussing the Generative Recommendation (GR) task it aims to solve.

\noindent \paragraph{Generative Recommendation.} Let $\mathcal U$ be a set of users interacting with a set of items $\mathcal I$. For each user $u\in \mathcal{U}$, we are given their interaction sequence of length $n_u$, denoted by $(i_1, \ldots, i_{n_u}) \in \mathcal{I}^{n_u}$. Our goal is to predict the next item $i_{n_u + 1}$ that user $u$ will interact with, given their past interactions. The GR framework addresses this task by modeling the probability distribution over interaction sequences and then outputting the most likely item according to $p(i_{n_u+1} \mid i_1, \ldots, i_{n_u})$.

% \paragraph{Generative Recommendation with Semantic IDs} The generative recommendation with the Semantic ID framework models the probability of sequences on the semantic IDs generated using features of each item instead of directly modelling on the item IDs. More specifically, GR with SID is typically divided into two steps: \textit{Semantic ID Generation (SID) step} and \textit{Generative Recommendation step} \cite{rajput2023recommender, deng2025onerec, hou2025generating}.  

% \vspace{4mm}

\noindent \paragraph{Generative Recommendation with Semantic IDs.} The GR with Semantic ID (GR with SIDs) paradigm represents items with tuples of SIDs derived from item semantic features, and models the probability of user sequences over SIDs, rather than directly over item IDs. More specifically, GR with SIDs consists of two steps: \textit{item SID assignment}  and  \textit{user sequence modeling} \cite{rajput2023recommender, deng2025onerec, hou2025generating}.

% In \textit{Semantic ID Generation (SID) step}, we generate a semantically meaningful sequence $(s_{i_j}^1, s_{i_j}^2, \ldots, s_{i_j}^m)$ for each item $i_j$ using information with item $i_j$. This is done by first obtaining an embedding $h_{i_j}$ of the item's semantic features with a pre-trained modality encoder, and then mapping the embeddings into the sequence of semantic IDs using residual quantization \cite{rajput2023recommender, deng2025onerec} or product quantization \cite{hou2025generating}. 

\textit{Item SID assignment.} The goal of {\em item SID assignment} is to represent items based on their rich semantic features, such as text and/or visual features, in a concise, scalable way. To do this, we first obtain a semantic embedding $h_{i_j}$ for each item $i_j \in \mathcal{I}$ by feeding  the item’s semantic features through a pre-trained semantic encoder, such as a large language or vision model.
Then, we execute a multi-layer clustering algorithm, such as Residual K-Means (RK-means) \cite{luo2024qarm, deng2025onerec}, Residual Quantized Variational Autoencoder (RQ-VAE) \cite{zeghidour2021soundstream, rajput2023recommender}, or product quantization \cite{opq, productquantization, hou2023learning}, on the set $\{h_{i_j}\}_{j \in \mathcal{I}}$ of all items' semantic embeddings. Suppose that we have $m$ clustering layers, then the cluster assignments $(s_{i_j}^1, s_{i_j}^2, \ldots, s_{i_j}^m)$ for each item $i_j$ form a semantically meaningful tuple. We treat the assignments at different layers as distinct tokens, i.e. if we have $c$ clusters per layer, then $s_{i_j}^1 \in \{1, \ldots, c\}, s_{i_j}^2 \in \{c+1, \ldots, 2c\}$ and so on. In this way, $(s_{i_j}^1, s_{i_j}^2, \ldots, s_{i_j}^m)$ forms the SID tuple for item $i_j$.

\textit{User sequence modeling.} The goal of \textit{user sequence modeling} is to model the probability of each user’s interaction sequence over the SIDs. In particular, for an interaction history $(i_1, \ldots, i_{n_u})$, GR with SIDs models the probability of the sequence $$S^u:=(s_{1}^1, \ldots , s_{1}^m, s_2^{1},  \ldots, s_2^m, \ldots, s_{n_u}^1, \ldots, s_{n_u}^m)$$ corresponding to the sequence of SIDs of the items interacted with by user $u$. 
% Once we have learned $ p(S_u)$, we can infer $(s_{n_u + 1}^1, \dots, s_{n_u}^m)$ by sampling from 
% $$p(s_{n_u + 1}^1, \dots, s_{n_u}^m | S_u) = p(S_u \cup (s_{n_u + 1}^1, \dots, s_{n_u+1}^m))/p(S_u),$$ where both terms in the RHS are likelihoods of user sequences and have been learned.
Existing methods primarily rely on autoregressive modeling, factorizing the probability either by each SID (eq.\eqref{eq:next-token-prediction}) or by each item’s full SID sequence (eq.\eqref{eq:next-item-prediction}): 
\begin{align}
    & p(S^u) = \prod_{i=1}^{n_u} \prod_{j=1}^m p( s_i^j \; | \; s_1^1,  \ldots, s_{i-1}^m, s_i^{< j} ) \label{eq:next-token-prediction} \\
    & p(S^u) =  \prod_{i=1}^{n_u} p( s_i^1 ,\ldots, s_i^{m} | s_1^1 ,\ldots, s_{i-1}^m ) \label{eq:next-item-prediction}
\end{align}
% Factorizing the probability by each SID (using \cref{eq:next-token-prediction}) is typically used with encoder-decoder or decoder only architecture and next-token prediction loss \cite{rajput2023recommender, deng2025onerec, grid}. \cite{hou2025generating} uses \cref{eq:next-item-prediction} by creating an embedding of the sequence using decoder transformer architecture and use it to predict all SIDs of the next item. The training of these methods are performed by maximizing the probability of the SID sequence on the training user-item interaction sequences. 
Factorizing by each SID (eq.\eqref{eq:next-token-prediction}) is the typical approach used in GR with SIDs methods \cite{rajput2023recommender, deng2025onerec, grid}. It is typically paired with encoder–decoder or decoder-only transformer architecture and a next-token prediction loss, making no assumptions on conditional independence of the tokens. On the other hand, \citet{hou2025generating} proposed eq.\eqref{eq:next-item-prediction} along with a method to embed the sequence using a decoder transformer and predict all SIDs of the next item jointly, assuming that the SIDs for the next item are conditionally independent given the previous SIDs. 
These methods require sequential inference of unseen SIDs, or in the case of \cite{hou2025generating}, unseen items, since they model the probability of unseen SIDs or items as dependent on all previous SIDs or items, including other unseen ones.
% {\color{red} -- please see Appendix \ref{app:formulation} for more details}. 
Further, the number of training targets is limited by the number of SIDs or items in the sequence (i.e. the number of factors in equations \eqref{eq:next-token-prediction} and \eqref{eq:next-item-prediction}, respectively).
% perform likelihood-based training over user–item interaction histories. 
To overcome these limitations, we propose to instead model the probability of SID sequences using masked diffusion.   

% Training in both cases is performed by maximizing the likelihood of SID sequences over user–item interaction histories.

% train the models using the next SID on the sequence \cite{rajput2023recommender, deng2025onerec} or by predicting all $m$ SIDs of the next item $\cite{hou2025generating}$ using an encoder-decoder model or decoder-only model. In this work, we 

\section{Proposed Method: MaskGR}

Recent works have demonstrated the power of masked diffusion models for modeling complex discrete distributions in NLP as well as protein design \cite{lou2024discrete, sahoo2024simple, shi2025simplified, ou2024absorbing}. Motivated by these results and the limitations of AR modeling, we propose \textbf{Mask}ed Diffusion over SIDs for \textbf{G}enerative \textbf{R}ecommendation, i.e. \textbf{MaskGR}, a framework that models the probability distribution of SID sequences through masked diffusion. 

Similar to prior works on GR with SIDs, we first generate a semantically meaningful SID tuple for each item and represent a user's interaction history by converting the item ID sequence $(i_1, \ldots, i_{n_u})$ into the corresponding SID sequence: $S^u := (s_{1}^1, \ldots, s_{1}^m, s_2^{1},  \ldots,$ $s_2^m, \ldots, s_{n_u}^1, \ldots, s_{n_u}^m)$. To introduce the masked diffusion framework over SIDs, we consider $n_u = n$ for all users $u\in\mathcal{U}$.

% In this section, we introduce our framework of modeling the probability of the SID sequence using the masked diffusion models. Similar to prior works on GR with Semantic IDs, we generate a semantically meaningful sequence of semantic IDs for each item and convert an interaction history from item IDs $(i_1, \ldots, i_{n_u})$ to the interaction history of SIDs $(s_{1}^1, \ldots , s_{1}^m, s_2^{1},  \ldots, s_2^m, \ldots, s_{n_u}^1, \ldots, s_{n_u}^m)$. We denote the SID sequence by $S^u$. To introduce the masked diffusion framework over SIDs, we consider all the SID sequences to be of length $n$. 

%Our method does not assume anything about the SID generation step and, therefore, can be combined with any SID generation step. 

\begin{figure*}
    \centering
    \includegraphics[width=0.4725\linewidth]{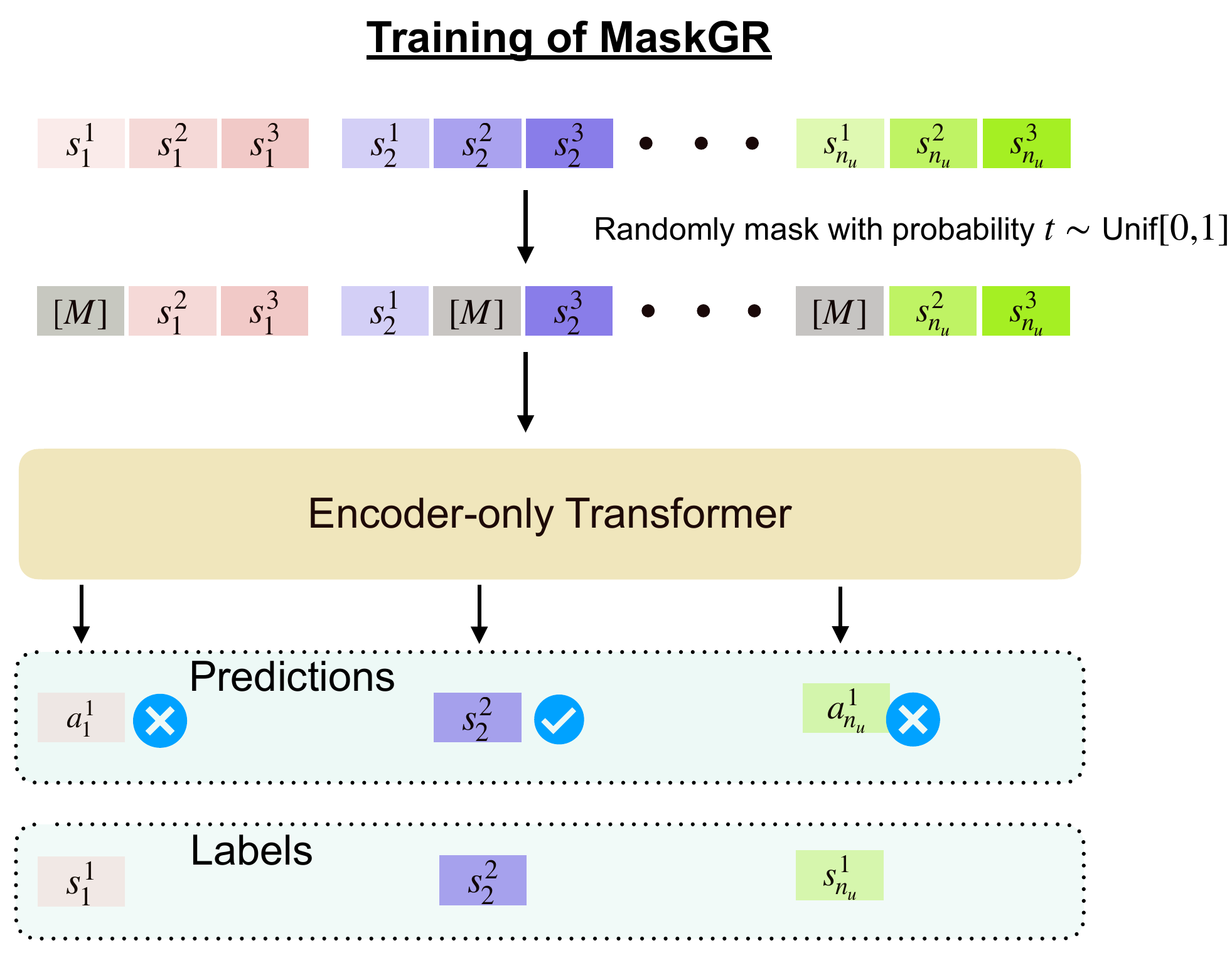}\hspace{5mm}
    \includegraphics[width=0.45\linewidth]{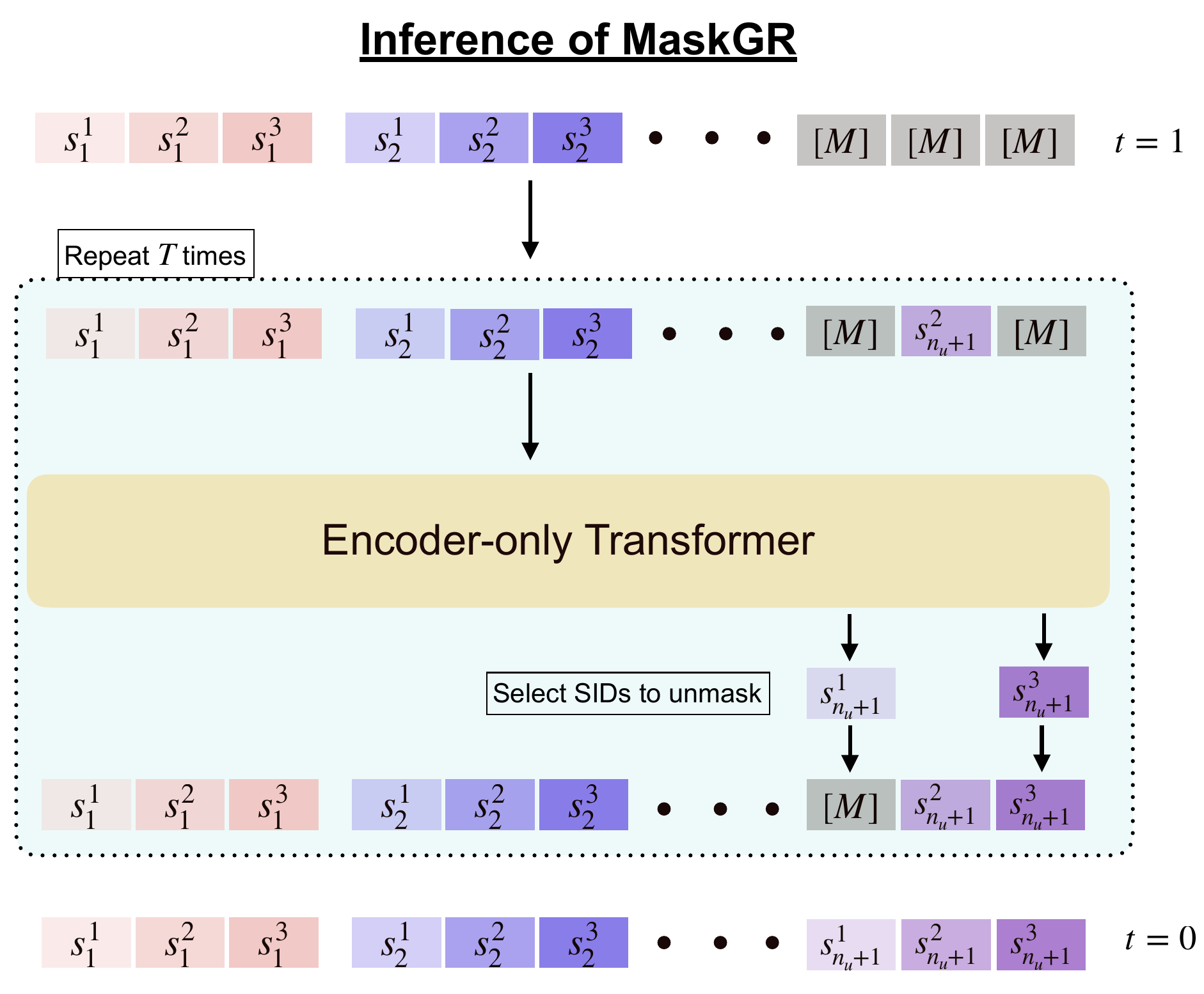}
    \caption{Overview of training and inference of MaskGR. During training, MaskGR randomly masks each SID in the SID sequence with a probability $t \sim \textrm{Unif}[0, 1]$ and feeds the masked sequence into an encoder-only transformer. The model is then optimized to reconstruct the original values of the masked SIDs using a cross-entropy loss applied at the masked positions (see Eq.~\eqref{eq:madrec-loss}). During inference, MaskGR begins with all SIDs of the last item replaced by masks. At each inference step, the partially masked sequence is passed through the network to predict values for all masked positions. The model then selectively unmasks a subset of these positions by retaining their predicted values while keeping the remaining positions masked. This iterative process continues until all SIDs are unmasked.}
    \label{fig:placeholder}
\end{figure*}

\subsection{MaskGR Training}

% [some context/motivation for why we use masked diffusion]
We model the probability of the SID sequence $S^u$ using discrete diffusion models with the masking noise framework \cite{lou2024discrete, sahoo2024simple, shi2025simplified, ou2024absorbing}. Before explaining the framework, we start with  notations. Let 
% $p_{\mathrm{SID}}$ denote the distribution over SID sequences and
$\mask$ denote the special mask token, $S^u_t$ denote the corrupted SID sequence at noise level $t \in [0, 1]$, and $S^u_t(i)$ be its $i^{\mathrm{th}}$ element. 
% Recall that $m$ is the number of SID tokens per tuple, and $c$ is the vocabulary size of each SID token.

\noindent \paragraph{Forward Process.} The forward process corrupts the original sequence, $S^u_0 = S^u$, by independently applying masking noise to each of its SID tokens. Specifically, each token in $S^u_0$ is replaced with the $\mask$ token with probability $t$, resulting in the noisy sequence $S^u_t$. Formally, the transition probability from the clean sequence $S^u_0$ to the noisy sequence $S^u_t$ is defined as:
% \noindent The forward process of masked diffusion models first samples $x_0$ from the data distribution $p_{\textrm data}$ and adds masking noise to each of its element independently by replacing them with $\mask$. In particular, the transition probability of $x_t$ given $x_0$ is given by
\begin{align*}
     &p(S^u_t | S^u_0) = \prod_{i=1}^{ m n } p( S^u_t (i) | S^u_0 (i) ), \\ 
    \quad \text{where} \;\; & p(  S^u_t (i) | S^u_0 (i)) = \text{Cat}( (1-t) e_{S^u_0 (i)} + t e_{\mask} ),
\end{align*}
where $\text{Cat}(\cdot)$ denotes the categorical distribution over $mc + 1$ possible tokens, and $e_x$ represents the one-hot vector corresponding to token $x$. Recall that $m$ denotes the number of SIDs per item and $c$ is the codebook size of each SID.

\noindent \paragraph{Reverse Process.}  To derive the denoising direction, we first compute the posterior $p(S^u_\ell | S_t^u, S_0^u)$ for some noise scale $\ell < t$. Since the forward process is coordinate-wise independent, the posterior can be coordinate-wise decomposed into $$p(S^u_\ell | S_t^u, S_0^u) = \prod_{i=1}^{m n} p(S^u_\ell (i) | S_t^u, S_0^u )$$ where the posterior for each coordinate is given by
\begin{align*}
    p(S^u_\ell (i) | S_t^u, S_0^u ) = \begin{cases}
        \text{Cat}( e_{S_t^u(i)} ) &\text{if} \;\; S_t^u(i) \neq \mask \\
        \text{Cat}( \frac{\ell}{t} e_{\mask} + (1 - \frac{\ell}{t}) e_{S_0^u(i)} )  &\text{if} \;\; S_t^u(i) = \mask 
    \end{cases}.
\end{align*}
% We approximate $e_{S_0^u(i)}$ by a neural network $f_\theta(\cdot | S_t^u)$ that predicts the probability distribution over possible SID values for each masked token. 

We approximate the above denoising posterior probability by approximating $e_{S_0^u(i)}$ with $f_\theta( \cdot | S_t^u): \{1, \dots, mc, \mask\}^n \rightarrow \mathcal{P}^{mc}$, where $\mathcal{P}^{mc}$ is the probability simplex over $mc$ discrete elements.  Specifically, the model $f_\theta$ takes an SID sequence $S_t^u$—in which some SIDs are masked—and outputs the predicted probability distribution over $mc$ elements for the possible values at each of those masked positions. In practice, $f_\theta$ is a neural network, typically a transformer encoder, with parameters $\theta$.

\noindent \paragraph{Training Objective.} The model parameters $\theta$ are trained by maximizing the evidence lower bound (ELBO) on the likelihood, which simplifies to a cross-entropy loss over the masked tokens:
\begin{align}
\label{eq:madrec-loss}
    L = \!\!\!\!\! \underset{\substack{t \sim \Unif[0, 1], \\ S^u_0 \sim p_{\mathrm{SID}}, \; S^u_t \sim p( S^u_t  | S^u_0 )}}{\E} \left[ - \frac{1}{t} \sum_{i = 1}^{m n} \mathbb{I}[ S^u_t(i) = \mask] \log p_{\theta} ( \; S^u_0(i) | S^u_t \; ) \right],
\end{align}
where $p_{\text{SID}}$ denotes the distribution over SID sequences, and the indicator function $\mathbb{I}[ S^u_t(i) = \mask]$ ensures the loss is computed only for masked SIDs.

This training objective randomly masks a subset of SIDs from the sequence according to the transition probability $p(S_t^u \mid S_0^u)$ and predicts the original values of the masked SIDs. This formulation is reminiscent of BERT4Rec \cite{sun2019bert4rec}, which randomly masks a fixed proportion of item IDs and learns to reconstruct them. In contrast, our approach follows the transition-based masking process, is mathematically grounded, and provably samples from the underlying probability distribution of the SID sequence \cite{lou2024discrete}. 

After training, MaskGR learns the conditional distribution of each masked SID given the remaining sequence. The next section describes the inference procedure and beam search for MaskGR. 

\subsection{MaskGR Inference and Beam Search} \label{madrec:inference}

To describe MaskGR inference, consider a user interaction history of item IDs $(i_1, \ldots, i_{n-1})$ with the corresponding SID sequence $Q = (s_1^1, \ldots, s_{n-1}^m)$, and assume we want to predict the next item $i_n$. To generate a sample using the learned denoising distribution $p_{\theta}(S_0^u(i) \mid S_t^u)$, MaskGR begins by masking the SIDs of the $n$th item, forming $\Tilde{S}_1 = (Q, A_1)$ where $A_1 = (\mask, \ldots, \mask)$. To transition from $\Tilde{S}_t = (Q, A_t)$ to $\Tilde{S}_r = (Q, A_r)$ for a lower noise scale $r < t$, the model iteratively unmasks a selected subset of masked tokens, sampling their values from the learned distribution $p_{\theta}(\Tilde{S}_r(i) \mid \Tilde{S}_t)$. 
% We note that changing the size of the subset of tokens to unmask trades off inference efficiency and performance, as we demonstrate in Section \ref{sec:inference-tradeoff}. Here, for simplicity, we consider unmasking one token per round.

In standard masked diffusion inference, each masked token is unmasked independently with probability $1 - (r/t)$. However, recent works \cite{zheng2024reparameterized, kim2025train, ben2025accelerated} demonstrate that performance can be substantially improved by selecting tokens to unmask based on the model’s prediction uncertainty at the masked positions. This unmasking process is repeated until all SIDs are unmasked.

% The set of tokens is chosen either randomly or using some predefined rule, such as uncertainty/confidence in the learned model $p_{\theta}(x_0^i | x_t)$ . 

The masked diffusion inference results in a \emph{sample} of an item from the probability conditioned on $Q$. In contrast, GR methods typically \emph{deterministically} output the most probable items given the interaction history by using beam search to find the most probable items. 

\noindent \paragraph{Beam Search in MaskGR.} Unlike AR models, which decode tokens in a fixed left-to-right order, MaskGR can generate SIDs in any order. Therefore, to understand the beam-search, suppose the tokens of the next item are unmasked following a specific order $(k_1, \ldots, k_m)$. Beam search in MaskGR aims to maximize the probability of the generated item by computing
\begin{align}
    p_{\theta}(s_n^{1}, \ldots, s_n^{m} \; | \; Q) =  \prod_{i=1}^m p_{\theta}( \; s_n^{k_i} \; | \; Q, s_n^{k_1}, \ldots, s_n^{k_{i-1}} ). 
\end{align}
Notably, the generation order does not need to be predetermined; it can be dynamically adjusted during inference. In our experiments, we evaluate three generation strategies: (1) random order as in vanilla masked diffusion models, (2) uncertainty-based order that prioritizes tokens with lower prediction uncertainty, and (3) left-to-right order. Please see Section~\ref{sec:inference-strategies} for more details.
% Observe that the SID generation order does not need to be predetermined before the start of the generation, and can be adapted during the inference. In our experiments, we experiment with three generation orders: random order given by vanilla MDM, an uncertainty based order and left-to-right order. For more discussion on the generation order, see Section~\ref{sec:inference-strategies}.

\noindent \paragraph{Beam Search in MaskGR with Multi-Token Prediction.} The MaskGR framework also supports predicting multiple tokens simultaneously, enabling the generation of the full $m$-length SID sequence with fewer than $m$ sequential function evaluations. Since MaskGR models only the conditional probabilities of masked tokens and not the joint probability among the masked tokens, i.e. it assumes the masked tokens are conditionally independent given the unmasked tokens, it \emph{approximates} the probability of the generated item $p_{\theta}(Q, s_n^{k_1}, \ldots, s_n^{k_m})$ during beam search.

To generate the $m$-length SID tuple of the next item, consider that $T < m$  model evaluations are used and in the $i$th step, $\alpha_i$ SIDs at positions $(k^i_{1}, \ldots, k^i_{\alpha_{i}} )$ are unmasked. For simplicity, assume the sequence of unmasking counts $(\alpha_1, \ldots, \alpha_T)$ are predetermined. In this case, we approximate the probability of the item as follows:

% Assume we want to generate the $m$-length SID of the next item using $\ell$ number of network evaluations, where $\alpha_i$ number of SIDs at positions $(k^i_{1}, \ldots, k^i_{\alpha_{i}} )$ are getting unmasked in $i^{\text{th}}$ steps. For simplicity, we assume that $(\alpha_1, \ldots, \alpha_\ell)$ are predetermined. In that case, we approximate the probability of the generation item as follows:
\begin{align*}
    p_{\theta}(s_n^{1}, \ldots, s_n^{m} | Q) = \prod_{j=1}^T \prod_{i=1}^{\alpha_j} p_{\theta}( \; s_n^{k_i^j} \; | \; Q, s_n^{k_1^1}, \ldots, s_n^{k^{j-1}_{\alpha_{j-1}}} ).
\end{align*}
In other words, we approximate the joint probability of all SIDs $(k^i_{1}, \ldots, k^i_{\alpha_{i}} )$ being unmasked at step $i$ by the product of their conditional probabilities. We use this approximation to guide the beam search inference in MaskGR while predicting multiple SIDs. Empirically,  we find that decoding multiple tokens simultaneously outperforms AR models despite this conditional independence assumption; please see Section \ref{sec:inference-tradeoff} for more details.
In the following section, we demonstrate how to extend MaskGR to incorporate an auxiliary method, namely fusing dense and generative retrieval, that was originally developed for AR-based SID modeling.

\subsection{Extending MaskGR with Dense Retrieval}
\label{sec:madrec-dense}

AR modeling is a dominant paradigm to model the SID sequences in GR. Several prior works have extended it, for instance, by combining SID generation with dense retrieval \cite{yang2024unifying, yang2025sparse}, incorporating user preferences \cite{paischer2024preference}, or developing methods for long SID generation \cite{hou2025generating}. MaskGR presents a novel approach to modeling the probability distribution of SID sequences. We believe that, in principle, MaskGR can be integrated with these existing AR extensions. To demonstrate this potential, we integrate MaskGR with dense retrieval, inspired by the work of Yang et al. \cite{yang2024unifying}.

To unify MaskGR's SID generation capabilities with dense retrieval, we modify the framework to output a dense item embedding in addition to the sequence of SIDs.  Specifically, we introduce three key modifications to the MaskGR framework:

\begin{itemize}[leftmargin=*]
    \item \textbf{Input Representation}: Instead of using only SID embeddings as input to the encoder-only architecture, we combine them with the text representation of each item. Formally, for an item $i$ with SID sequence $(\sigma_i^1, \ldots, \sigma_i^m)$, the input embedding is defined as:
    \begin{align*}
        H_{\sigma_i^j} = h_{\sigma_i^j} + A_j h^{\textrm{text}}_i,
    \end{align*}
    where $h_{\sigma_i^j}$ is the embedding of the SID token $\sigma_i^j$, $h^{\textrm{text}}_i$ is the embedding of item $i$'s text features extracted from a language model, and $A_j$ is a learnable linear transformation that projects the text embedding to the same dimension as the SID embeddings.
    \item \textbf{Masking Strategy}: In the original MaskGR, each SID is independently masked with a probability proportional to the noise scale $t$. However, since the dense embedding corresponds to the item as a whole, we encourage learning at the item abstraction level by masking all SIDs of an item jointly. Concretely, during training, with a fixed probability $\beta$, we mask all SIDs of an item together, using the same noise scale as the masking probability, and with probability $1 - \beta$, we use the masking strategy of the original MaskGR.  Here, $\beta$ is a hyperparameter. 
    
    \item \textbf{Prediction Mechanism and Loss Function}: After a fixed number $\eta$ of layers in the network $f_{\theta}$, we use the resulting hidden states to form the predicted dense embedding. More precisely, to predict a dense embedding of an item, we mask all its SIDs and  pass the full sequence through $f_{\theta}$, as described in Section \ref{madrec:inference}. Let $\Tilde{h}^j$ denote the embedding of the $j$-th SID after $\eta$ layers; we concatenate the predicted embeddings for all $m$ SIDs to obtain the predicted dense embedding $\Tilde{H} = \{ \Tilde{h}^1, \ldots, \Tilde{h}^m \}$. Next, we project the predicted dense embedding to the same dimension as the text item embedding with a linear or a small MLP $g_{\theta}$. The dense retrieval objective encourages $\Tilde{E}$ to align with the ground-truth text embedding of the corresponding item, formulated as:  
    \begin{align*}
        \mathcal{L}_{\textrm{dense}} = - \log \frac{ \exp\!\left( g_{\theta}(\Tilde{H})^\top h_i^{\textrm{text}} \right) }{ \sum_{j \in \mathcal{I}} \exp\!\left( g_{\theta}(\Tilde{H})^\top h_j^{\textrm{text}} \right) },
    \end{align*}
    where $\mathcal{I}$ denotes the set of all items.  The dense retrieval loss is applied only to items whose SIDs are all masked.
\end{itemize}

These modifications extend MaskGR to unify SID generation with dense retrieval. We now turn to experiments to examine the performance of MaskGR. 

\section{Experiments}

In this section, we empirically study the following questions:
\begin{itemize}[leftmargin=*]
    \item \textbf{Q1.} How does the {\em overall performance} of MaskGR compare to AR modeling with SIDs and other GR baselines?
    \item \textbf{Q2.} How does MaskGR's perform in  {\em data-constrained settings}?
    \item \textbf{Q3.} How does MaskGR {\em trade off inference efficiency and performance}?
    \item \textbf{Q4.} How does MaskGR depend on its {\em components}?
    \item \textbf{Q5.} Does {\em extending  MaskGR with dense retrieval} improve its performance?
\end{itemize}

% provide the experiments to show the effectiveness of MaskGR and understand different aspects of the performance. 
% We start by describing our experimental setup. 

\subsection{Experimental Setup}

We first describe the experimental settings. Please see Appendix \ref{app:experiments} for more details.

\noindent \paragraph{Implementation details.} We use an 8-layer encoder-only transformer model with a 128-dimensional embedding and rotary position embedding. The model includes 8 attention heads and a multi-layer perceptron (MLP) with a hidden layer size of 3072. The total number of parameters in our model is 7M. Following \cite{grid}, we assign SIDs by first  extract 4096-dimensional text embeddings from item metadata with Flan-T5-XXL \cite{chung2024scaling}, then apply residual k-means clustering with four layers, each having a codebook size of 256. We append de-duplication token to distinguish items with identical SID tuples as in \cite{rajput2023recommender}. Unless specified otherwise, the number of inference steps  is equal to the number of SIDs per item. We use a greedy inference strategy to choose the set of tokens to unmask based on their prediction uncertainty, measured as the difference between the probabilities of the first and second most likely assignment at each masked SID position -- please see Section \ref{sec:inference-strategies} for more details.  

\noindent \paragraph{Evaluation.} We evaluate all methods using Recall@K and NDCG@K metrics, with $K \in \{5, 10\}$. We use the standard leave-one-out evaluation protocol, where the last item of each user's sequence is used for testing, the second-to-last for validation, and the remaining items for training \cite{kang2018self, CIKM2020-S3Rec, geng2022recommendation, rajput2023recommender, yang2024unifying}. 

\noindent \paragraph{Baselines.} We compare the performance of MaskGR against six representative baselines \cite{kang2018self, sun2019bert4rec, yang2023generate, cui2024cadirec, rajput2023recommender, yang2024unifying}:
\begin{itemize}[leftmargin=*]
    \item \textbf{Item ID-based methods:} SASRec \cite{kang2018self} and BERT4Rec \cite{sun2019bert4rec}, two widely used sequence recommendation models that operate directly on item IDs.  
    \item \textbf{Diffusion-based methods:} DreamRec \cite{yang2023generate} and CaDiRec \cite{cui2024cadirec}, which apply continuous diffusion processes on item IDs for recommendation. Both methods use 1,000 inference steps.
    \item \textbf{Generative recommendation with SIDs:} TIGER \cite{rajput2023recommender}, which autoregressively models the probability distribution over the SID sequence, and LIGER \cite{yang2024unifying}, which extends TIGER by integrating dense retrieval. 
\end{itemize}
% \cite{kang2018self, sun2019bert4rec, yang2023generate, cui2024cadirec, rajput2023recommender, yang2024unifying}. In particular, we use two popular item ID based recommendation baselines -- SASRec \cite{kang2018self} and BERT4Rec \cite{sun2019bert4rec}, two diffusion-based baselines - DreamRec \cite{yang2023generate} and CadiRec \cite{cui2024cadirec} -  and two GR with semantic IDs baselines -- TIGER \cite{rajput2023recommender} and LIGER \cite{yang2024unifying}. 

\noindent \paragraph{Datasets.} We evaluate our method on four public benchmarks: three categories (Beauty, Sports, and Toys) from the Amazon Review dataset \cite{mcauley2015image} and the MovieLens-1M (ML-1M) movie rating dataset \cite{Harper2016TheMD}. For the Amazon datasets we apply the standard 5-core filtering to remove users and items with fewer than 5 interactions, and use title, category, description and price as the item text features, following \cite{rajput2023recommender}. We also apply 5-core filtering to ML-1M, and use title, year and genres as the item text features. Table \ref{tab:data} provides statistics of the processed datasets.

\subsection{Q1. Overall Performance}

We begin by presenting our experimental results on the generative recommendation (GR) task, comparing MaskGR's performance with other GR methods across the four aforementioned datasets. The results, shown in Table~\ref{tab:main-result-horizontal-clean}, indicate that our proposed framework performs comparably or better than other GR methods.

\noindent \paragraph{Comparison with AR modeling.} The results indicate a {\em significant improvement over TIGER}, with an average of a 21.9\% increase in NDCG@5 across all datasets. Recall that TIGER autoregressively  models  the probability of the SID sequence. Our results support prior work showing that masked diffusion frameworks are more effective than autoregressive (AR) models in data-sparse settings like sequential recommendation \cite{prabhudesai2025diffusion, ni2025difflm}. 

\begin{figure}
    \centering
    \includegraphics[width=0.35\linewidth]{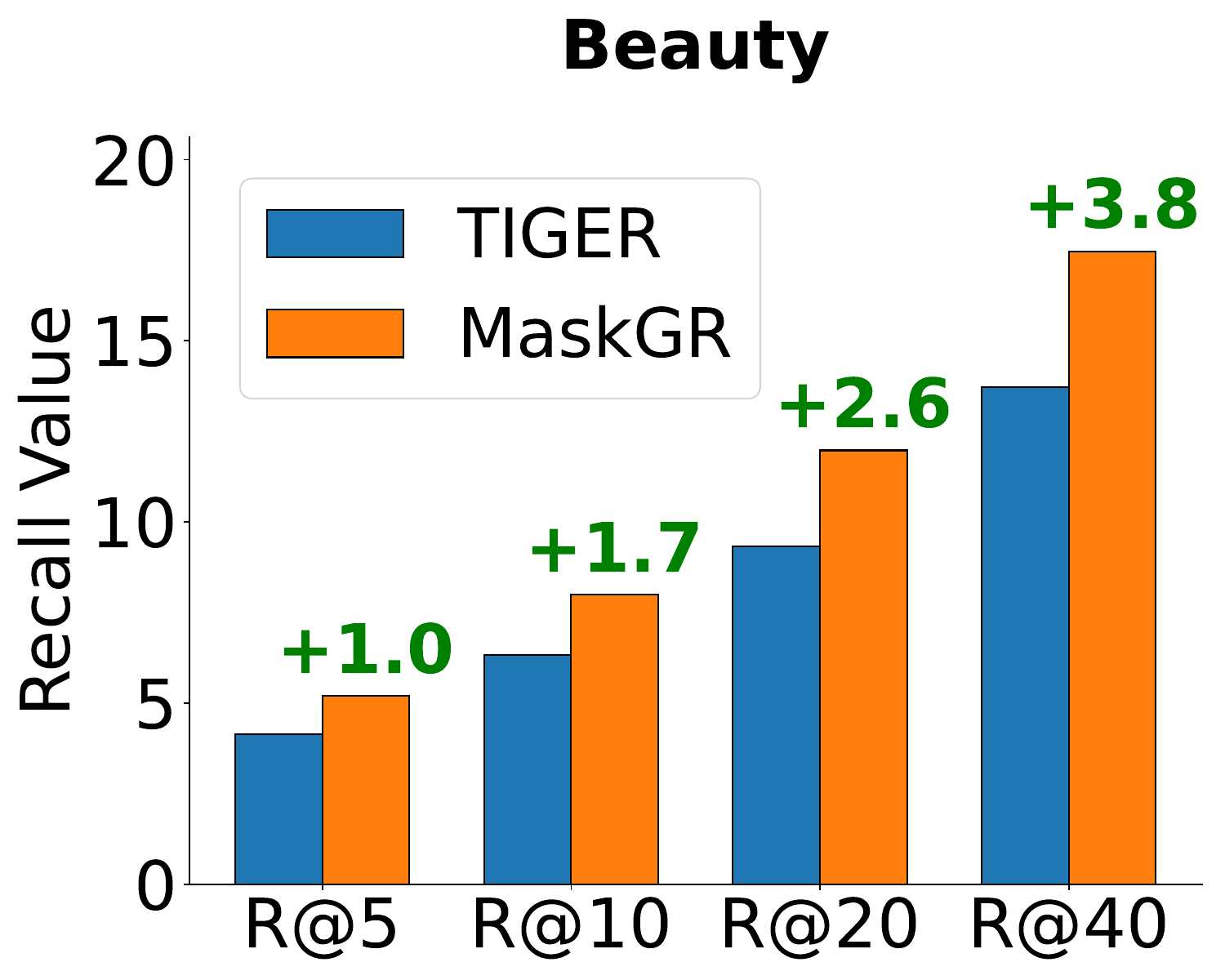}\hspace{7mm}
    \includegraphics[width=0.35\linewidth]{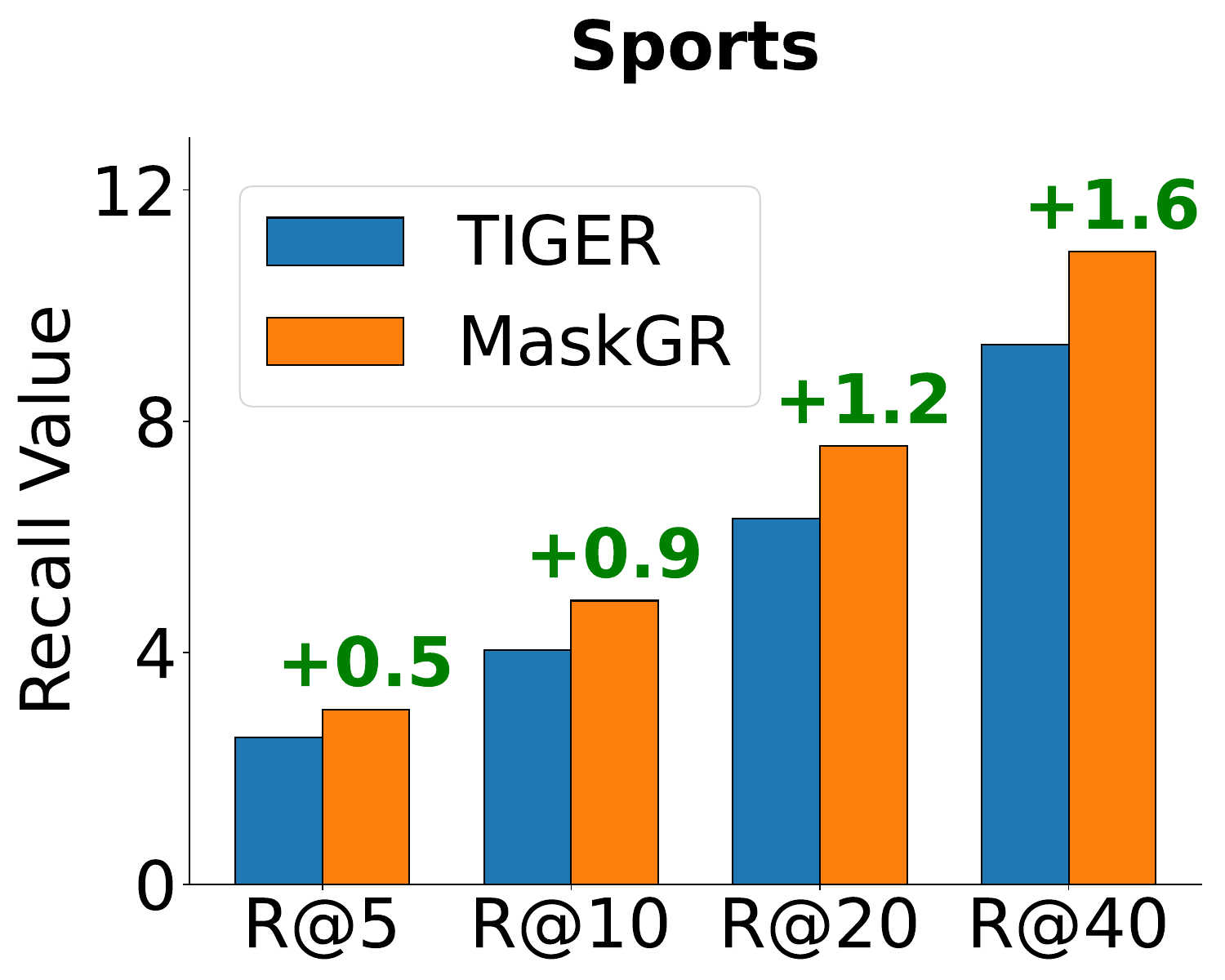}
    \caption{Improved performance gap for coarse-grained retrieval on the Beauty and Sports datasets. The gap in Recall@K between TIGER and MaskGR increases as K increases.}
    \label{fig:coarse-grained-perf}
\end{figure}

Moreover, we further observe in Figure~\ref{fig:coarse-grained-perf} that the performance gap between MaskGR and TIGER increases as we {\em increase the retrieval granularity}. Here we measure Recall@K for $K \in \{ 5, 10, 20, 40 \}$ on the Beauty and Sports datasets, and notice MaskGR's increasing performance gains with larger $K$, i.e., more coarse-grained retrieval. This suggests that masked diffusion is better able to model global relationships among tokens than AR modeling, as MaskGR's rankings tend to have higher-quality depth than TIGER's. This is consistent with prior results suggesting that AR modeling over-indexes on local relationships among tokens \cite{khandelwal2018sharp, henighan2020scaling, brunato2023coherent}, as well as our intuition that masked diffusion, by training a more diverse set of targets, is better able to model global relationships.

\noindent \paragraph{Comparison with continuous diffusion.} MaskGR also {\em outperforms continuous diffusion-based recommendation} approaches, namely DreamRec and CaDiRec, as shown in Table~\ref{tab:main-result-horizontal-clean}. Importantly, both DreamRec and CaDiRec require 1000 diffusion inference steps, whereas our method uses only 5 inference steps. This further supports the conclusion of prior works that diffusion with masking noise is more effective for generation in discrete domains than diffusion with Gaussian noise \cite{lou2024discrete, gong2024scaling}.

% \noindent \textbf{Improved Performance Gap for Coarse-Grained Retrieval.} We further observe in Figure~\ref{fig:coarse-grained-perf} that 

% To evaluate the performance of MaskGR on coarse-grained retrieval, we measure Recall@K for $K \in \{ 5, 10, 20, 40 \}$ on Beauty and Sports datasets. We measure the performance of MaskGR that models the probability of SID sequences using masked diffusion to TIGER that models the probability using autoregressive modelling. We report our results in Figure~\ref{fig:coarse-grained-perf}. We observe that the performance gap between MaskGR and TIGER increases as we increase $K$. This shows the improved gap between coarse-grained retrieval between MaskGR and TIGER.

% \kulin{Add hierarchy-wise accuracy}

\begin{table*}[htb]
    \centering
    % \scriptsize
    \caption{Performance comparison of MaskGR with other GR methods on the Beauty, Sports, Toys, and MovieLens-1M datasets. The best result for each metric is in  bold and the second best is underlined.
    % ${}^{1}$LIGER combines the autoregressive training of TIGER with dense retrieval.
    }
    \label{tab:main-result-horizontal-clean}
    \begin{tabular}{l *{8}{c}} 
        \toprule
        \multirow{2}{*}{\textbf{Method}} & \multicolumn{2}{c}{\textbf{Beauty}} & \multicolumn{2}{c}{\textbf{Sports}} & \multicolumn{2}{c}{\textbf{Toys}} & \multicolumn{2}{c}{\textbf{ML-1M}} \\
        \cmidrule(lr){2-3} \cmidrule(lr){4-5} \cmidrule(lr){6-7} \cmidrule(lr){8-9}
        & \textbf{R@5} & \textbf{N@5} & \textbf{R@5} & \textbf{N@5} & \textbf{R@5} & \textbf{N@5} & \textbf{R@5} & \textbf{N@5} \\
        \midrule
        SASRec & 3.87 & 2.49 & 2.33 & 1.54 & 4.63 & 3.06 & 9.38 & 5.31 \\
        BERT4Rec & 3.60 & 2.16 & 2.17 & 1.43 & 4.61 & 3.11 & 13.63 & 8.89 \\
        DreamRec & 4.40 & 2.74 & 2.48 & 1.51 & 4.97 & 3.16 & 13.04 & 8.58 \\
        CaDiRec & \underline{4.95} & 3.14 & \underline{2.76} & \underline{1.83} & \underline{5.22} & \underline{3.56} & \underline{15.04} & \underline{10.01} \\
        TIGER & 4.29 & 2.88 & 2.45 & 1.64 & 4.42 & 2.91 & 12.83 & 8.85 \\
        LIGER & 4.62 & \underline{3.17} & 2.61 & 1.67 & 4.66 & 3.01 & 13.73 & 9.12 \\
        \midrule 
        % \textbf{MaskGR} & \textbf{5.38} & \textbf{3.51} & \textbf{3.02} & \textbf{1.91} & \textbf{5.48} & \textbf{3.75} & \textbf{16.72} & \textbf{11.12} \\
        % \midrule
        % \textbf{Improv. (\%)} & +8.7 & +10.7 & +9.4 & +4.4 & +5.0 & +5.3 & +11.17 & +11.1 \\
        \textbf{MaskGR} & \textbf{5.38} & \textbf{3.51} & \textbf{3.02} & \textbf{1.91} & \textbf{5.48}  & \textbf{3.75}  & \textbf{16.72} & \textbf{11.12}  \\
        + Improv \% & +8.7 \% & +10.7 \% & +9.4 \% & +4.4 \% & + 5.0 \% & +5.3 \% & +11.2 \% & +11.1 \% \\
        % \midrule
        % \textbf{Improv. (\%)} &  & +10.7 & +9.4 & +4.4 & +5.0 & +5.3 & +11.2 & +11.1 \\
        \bottomrule
    \end{tabular}
\end{table*}

\subsection{Q2. Data-constrained Performance}

\begin{figure}
    \centering
    \includegraphics[width=0.35\linewidth]{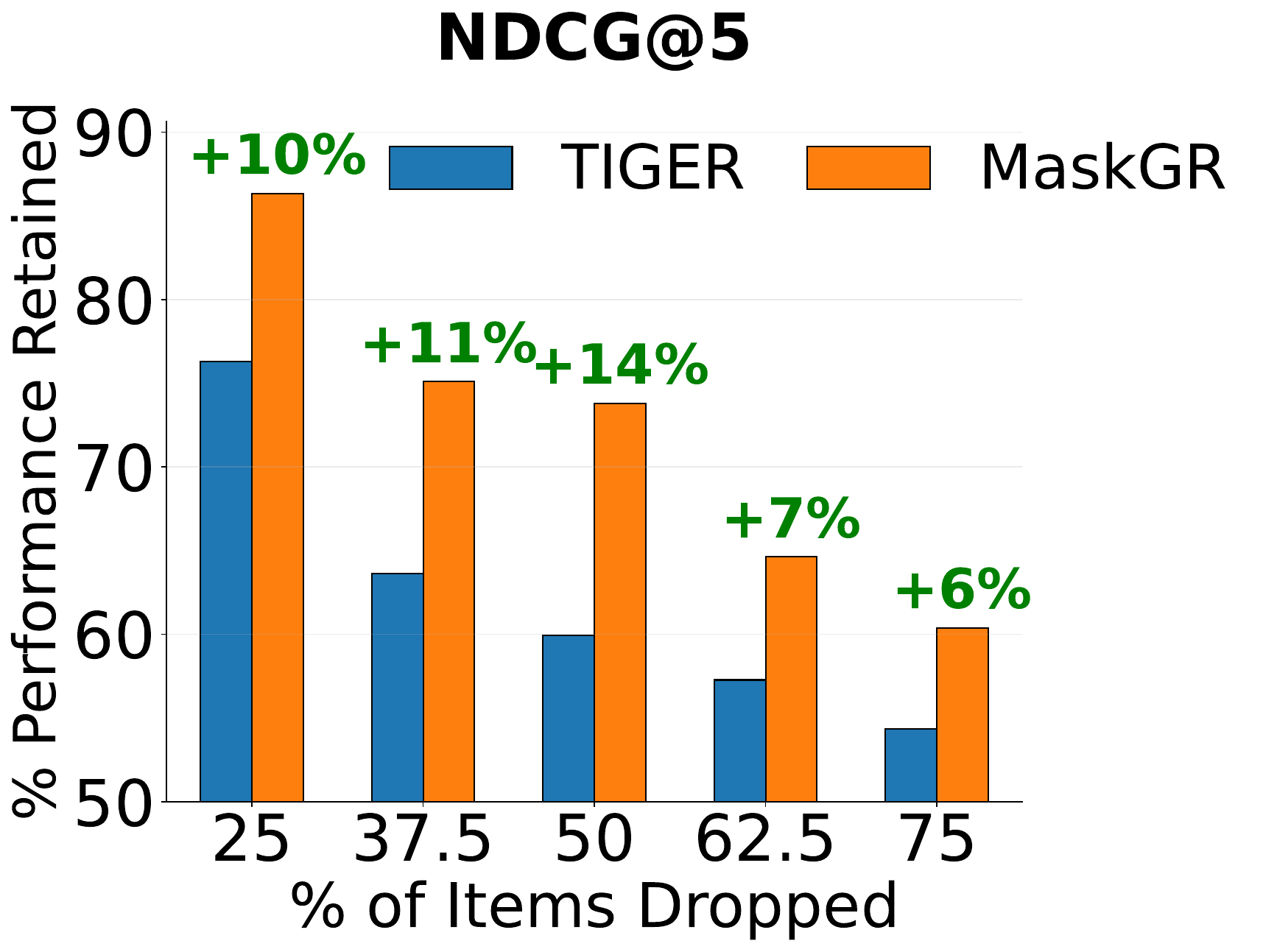}
    \hspace{7mm}
    \includegraphics[width=0.35\linewidth]{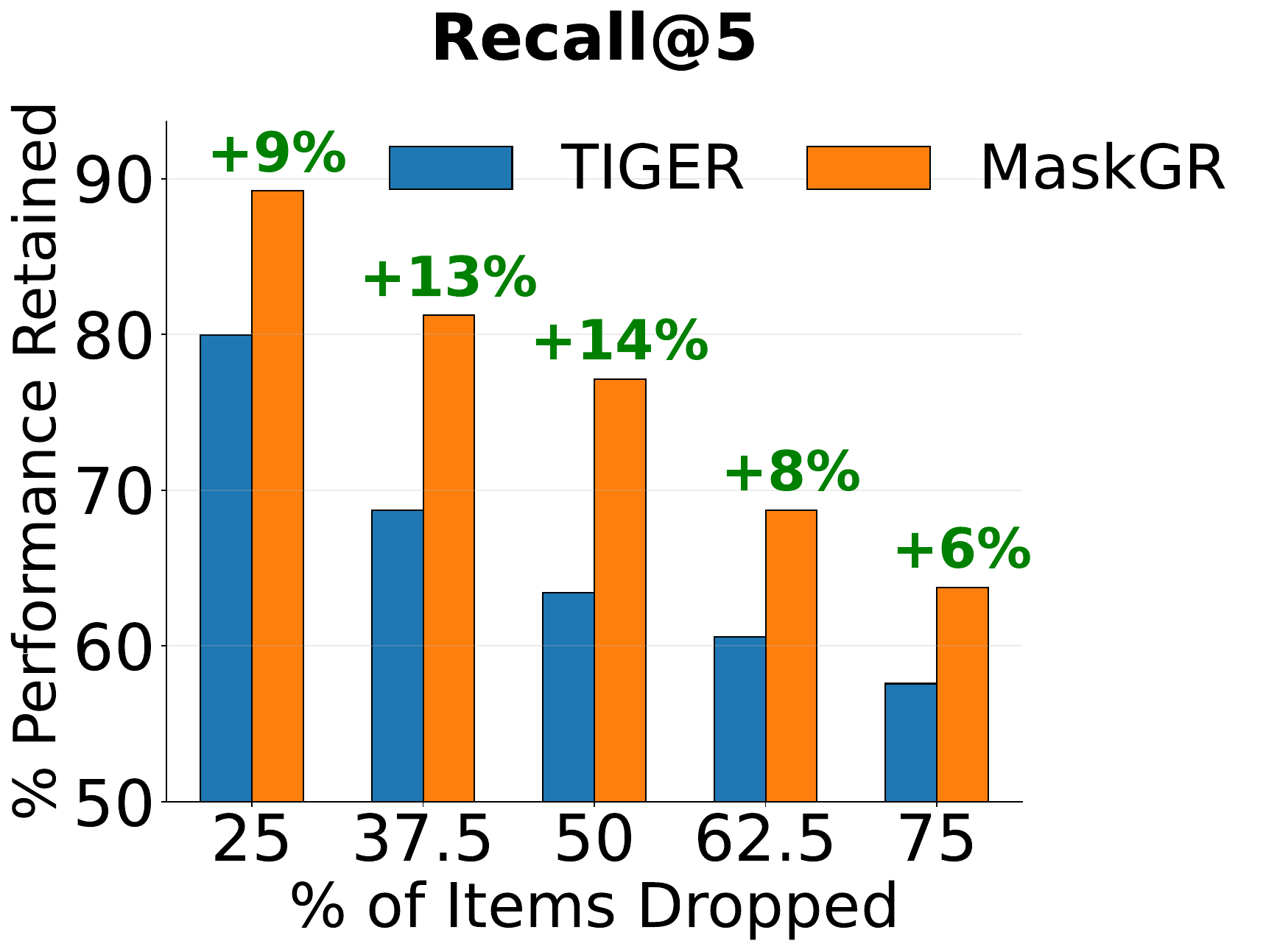}
    \caption{Comparison of data efficiency of MaskGR and TIGER by dropping 25\%, 37.5\%, 50\%, 67.5\% and 75\% of items from each sequence in the training set, while maintaining at least three items in each sequence.}
    \label{fig:data-constrained-fig}
\end{figure}

To better understand MaskGR's effectiveness in data-constrained settings, we evaluate its and TIGER's performance on increasingly sparsified versions of the Beauty dataset. Specifically, we drop 25\%, 37.5\%, 50\%, 62.5\%, and 75\% of items from each sequence in the training set, while ensuring each sequence has at least three items. The validation and test sets remain unchanged. We then measure the retained performance by comparing the Recall@5 and NDCG@5 scores of these models with the scores of a model trained without dropping items. The results are shown in Figure~\ref{fig:data-constrained-fig}. 

As the percentage of dropped items increases, TIGER’s performance drops faster than MaskGR's. At significantly high drop percentages (62.5\% and 75\% of items dropped), the performance drop for both methods becomes similar, but this is expected, as dropping close to 100\% of the items would result in near-zero performance retention for any method. 

\subsection{Q3. Inference Performance-Efficiency Trade-off} \label{sec:inference-tradeoff}

\begin{figure}
    \centering
    \includegraphics[width=0.35\linewidth]{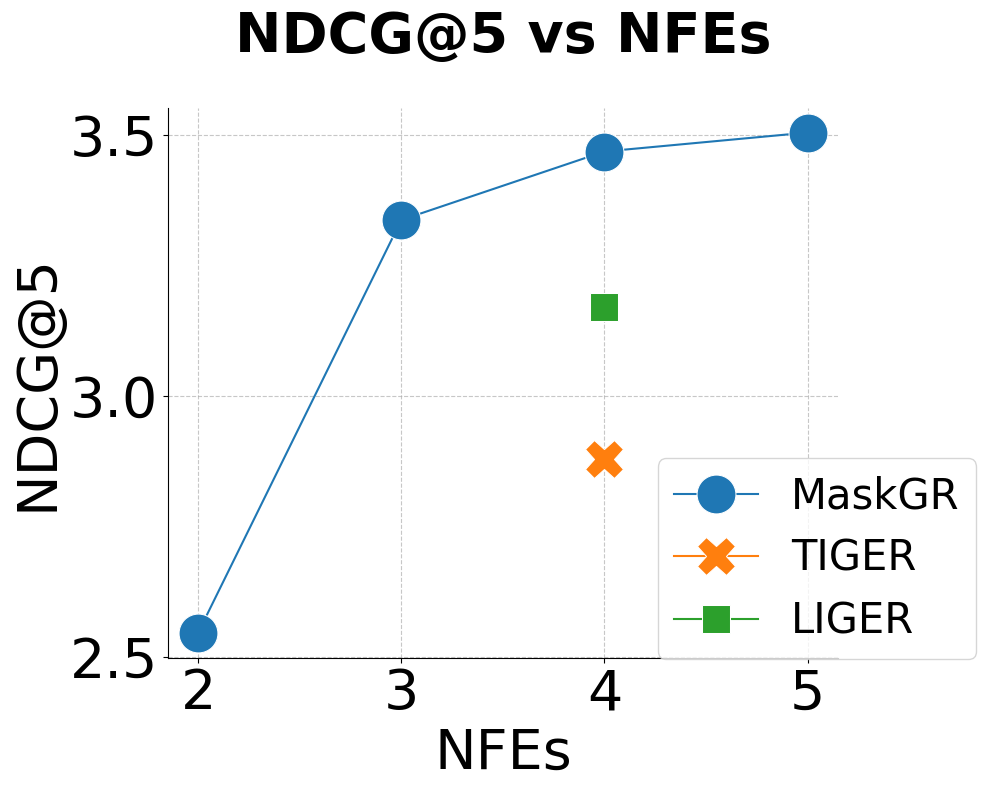}
    \hspace{7mm}
    \includegraphics[width=0.35\linewidth]{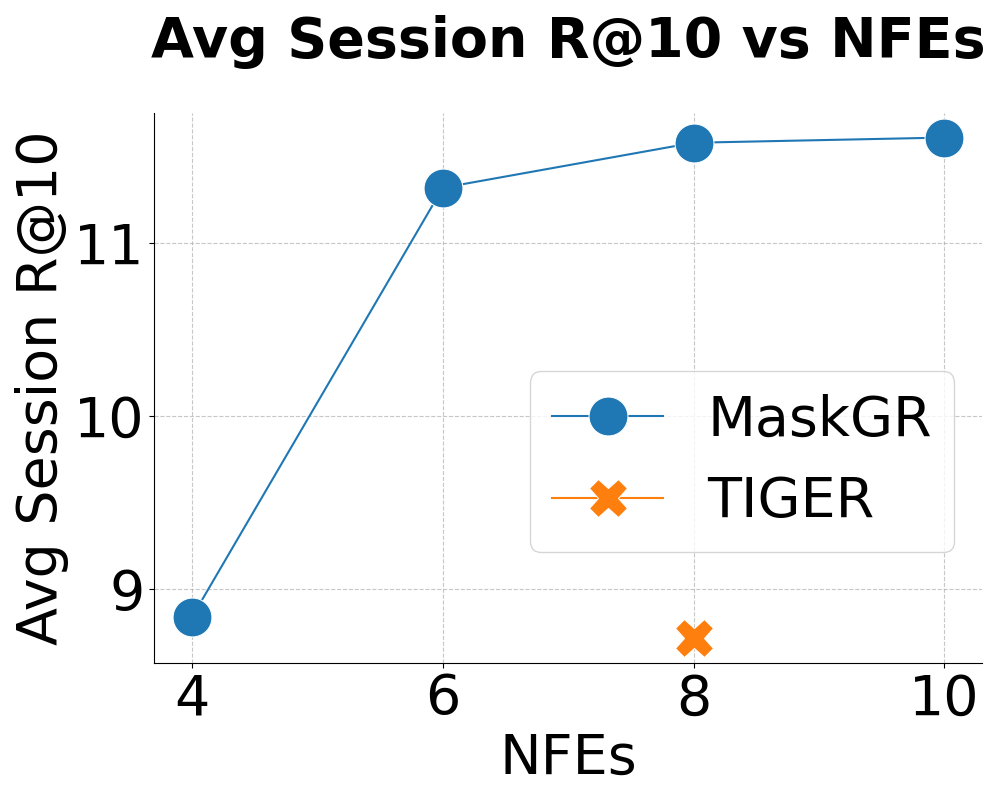}
    \caption{Next-$k$ item prediction performance vs number of function evaluations (NFEs) during inference for (Left) $k=1$ on  Beauty and (Right) $k=2$ on MovieLens-1M.
    The AR methods (TIGER and LIGER) must decode tokens sequentially, so they always execute $k \; \times $  (\# SIDS/item) NFEs. MaskGR can decode multiple items in parallel, thereby allows trading off performance and efficiency by tuning the NFEs. 
    Moreover, it already  outperforms the AR methods with fewer NFEs.
    }
    \label{fig:ndcg-nfes-plot}
\end{figure}

\noindent \paragraph{Single-item prediction.} Unlike AR methods, 
% MaskGR allows decoding multiple tokens simultaneously during inference without requiring any changes to the training procedure. This ability enables 
MaskGR can decode $m$ SIDs of an item with fewer than $m$ sequential function evaluations, albeit with a potential drop in performance. 
% As expected, however, this efficiency comes with a modest drop in performance. 
To study the trade-off between the number of function evaluations (NFEs, denoted as $T$ in Section \ref{madrec:inference}) and performance, we evaluate MaskGR with $2, 3, 4$ and $5$ NFEs on the Beauty dataset (MaskGR uses 5 SIDs per item). We compare the performance using NDCG@5 with AR baselines in Figure~\ref{fig:ndcg-nfes-plot}. Notably, even with only 3 NFEs, MaskGR surpasses the performance of TIGER by 13\% and LIGER by 4.7\%, which both use 4 NFEs. 

% Autoregressive methods such as TIGER, LIGER decode each SID sequentially. However, unlike the autoregressive methods, MaskGR provides an inference time flexibility of decoding multiple tokens without modifying the training. This ability helps MaskGR to decode $m$ SID of an item with less than $m$ (sequential) number of function evaluations but as expected this efficiency comes with some performance drop. To understand the tradeoff between number of function evaluations and the performance of MARec, we evaluate the performance with $2, 3, 4$ and $5$ sequential forward pass on Amazon Beauty dataset (recall MaskGR by default uses 5 number of SIDs). We compare our results with autoregressive methods such as TIGER and LIGER in Figure~\ref{fig:ndcg-nfes-plot}. We observe that even with 3 NFEs MaskGR outperforms TIGER and LIGER's performance that uses 4 NFEs. 

\noindent \paragraph{Muli-item prediction.} Intuitively, the benefit of MaskGR’s multi-token prediction should become more apparent as the number of items to recommend grows. To evaluate this hypothesis, we adapt the standard leave-one-out evaluation protocol into a leave-two-out protocol: the last two items of each user’s sequence are used for testing, the two preceding items for validation, and the remaining sequence for training. This evaluation protocol also aligns closely with session-wise recommendation tasks \cite{deng2025onerec}.

We train and evaluate MaskGR and TIGER on the ML-1M dataset using this leave-two-out protocol, choosing ML-1M because of its longer average sequence length. We fix the total number of generated beams to 10. In this setting, each beam represents a pair of items, so the number of candidate recommendations at each position is fewer than 10. After generating 10 session beams, we compute the recall for each predicted item and average the results, which we refer to as the \emph{average session recall@10}. Figure~\ref{fig:ndcg-nfes-plot} plots this metric against the number of function evaluations. We observe that MaskGR achieves the same performance as TIGER with only 4 NFEs—reducing NFEs by 50\% compared to TIGER’s 8. 

% \begin{figure}
%     \centering
    
%     \caption{Caption}
%     \label{fig:placeholder}
% \end{figure}

\subsection{Q4. Component-wise importance.}

We next execute several ablation studies to understand the importance of key components of MaskGR.

\noindent \paragraph{Importance of {\em semantic} IDs.}
To understand MaskGR’s ability to utilize semantic information from the SIDs derived from the item text embeddings, we conduct two complementary experiments on the Beauty dataset in which the provided semantic information is systematically removed. In the first experiment, instead of using SIDs generated from item embeddings, we replace them with randomly assigned tuples of tokens with the same vocabulary as the original SIDs. We refer to this method as MaskGR + Random SIDs. In the second experiment, instead of modeling the probability distribution over the SID sequence as in MaskGR, we directly model the probability distribution over the item IDs. More specifically, we model the item ID sequence $(i_1, \ldots, i_n)$ according to the framework of MaskGR in place of the SID sequence $(s_1^1, \ldots, s_n^m)$. We refer to this method as MaskGR + Item IDs. We report and compare the performance in Table~\ref{tab:sid-generation}.

We find that replacing the true SIDs computed by residual k-means clustering of semantic item embeddings with random SIDs reduces Recall@10 from 8.15 to 5.53, a drastic reduction. Moreover, using MaskGR on item IDs instead of SIDs decreases performance from 8.15 to 6.71. These results demonstrate that MaskGR effectively leverages the semantic information contained in the SIDs. 

\noindent \paragraph{Importance of dynamic masking probability.} In Table~\ref{tab:sid-generation}, we also compare against BERT4Rec \cite{sun2019bert4rec}. Interestingly, \emph{MaskGR trained directly on item IDs still outperforms BERT4Rec.}  BERT4Rec predicts randomly masked item IDs in the interaction history but masks only a fixed fraction of items during training ($t=0.15$), whereas MaskGR + Item IDs masks all possible fractions in the interval $[0, 1]$, enabling a more effective training regime.

% We observe that when we replace SIDs generated by RK-means clustering of the item embeddings in MaskGR to the random SIDs, the Recall@10 drops from 8.15 to 5.53. We also observe that when MaskGR is used on item IDs instead of semantic IDs, the performance drops from 8.15 to 6.71. These two experiment show that MaskGR is able to effectively leverage semantic information from the SIDs. It is also interesting to note that MaskGR on item IDs outperform BERT4Rec. Recall that BERT4Rec is also trained to predict the randomly masked item IDs in the interaction history but BERT4Rec only masks a fixed fraction of item IDs during the training while MaskGR masks all possible fractions in $[0, 1]$ during the training. 

\begin{table}[t]
    \centering
    % \small % optional: makes the table a bit more compact
    \setlength{\tabcolsep}{12pt} % adjust column spacing
    \caption{Comparison of Recall@K (R@K) and NDCG@K (N@K) for $K\in \{5, 10\}$ for different versions of MaskGR on the Beauty dataset. Note that BERT4Rec is effectively MaskGR + Item IDs with fixed masking ratio.}
    \begin{tabular}{lcccc}
        \toprule
        \textbf{Method} & \textbf{R@5} & \textbf{R@10} & \textbf{N@5} & \textbf{N@10} \\
        \midrule
        BERT4Rec & 3.60 & 6.01 & 2.16 & 3.08 \\
        MaskGR w/ Item IDs  & 4.69 & 6.71 & 3.12 & 3.77 \\
        MaskGR w/ Random SIDs  & 3.78 & 5.53 & 2.61 & 3.05 \\
        MaskGR & \textbf{5.38} & \textbf{8.15} & \textbf{3.51} & \textbf{4.41} \\
        \bottomrule
    \end{tabular}
    \label{tab:sid-generation}
\end{table}

\noindent \paragraph{Dependence on the number of SIDs per item.} To understand the performance of MaskGR as we scale number of SIDs per item, we perform an ablation on changing number of SIDs while fixing the rest of the setting. We report our results in Table~\ref{tab:diff-sid}. We observe  that the performance of MaskGR improves as we increase number of SIDs from 3 to 4 but decreases as we go from 4 to 5. A plausible reason for the decrease in performance from 4 to 5 is the potential increase in invalid predicted SIDs, as the model may predict SIDs that do not correspond to a valid item. Combining MaskGR with constrained beam search to prevent such invalid SIDs as we scale the number of SIDs would be an interesting future direction \cite{hou2025generating}. 

\begin{table}[h]
    \centering
    \setlength{\tabcolsep}{12pt}
    \caption{Performance comparison on the Beauty dataset as we scale the number of SIDs per item.}
    \label{tab:diff-sid}
    \begin{tabular}{c|cccc}
    \toprule
    \textbf{Number of SIDs} & \textbf{R@5} & \textbf{R@10} & \textbf{N@5} & \textbf{N@10} \\
    \midrule
    3 & 4.96 & 7.93 & 3.24 & 4.20 \\
    4 & 5.38 & 8.15 & 3.51 & 4.41 \\
    5 & 4.86 & 7.53 & 3.26 & 4.11 \\
    \bottomrule
    \end{tabular}
\end{table}

\noindent \paragraph{Role of inference strategy.}
\label{sec:inference-strategies}
As MaskGR training does not incorporate an inductive bias toward any particular token order -- such as the left-to-right ordering used in autoregressive training -- it offers greater flexibility in choosing which tokens to unmask during inference \cite{zheng2024reparameterized, kim2025train}. In this section, we evaluate three strategies for selecting the tokens to unmask:

\begin{itemize}[leftmargin=*]
    \item \textbf{Random inference}: follows vanilla masked diffusion modeling  inference in randomly selecting SIDs to unmask. 
    \item \textbf{Greedy inference}: chooses the SIDs based on their prediction uncertainty, measured as the difference between the probabilities of the first and second most likely assignments at each masked SID position. 
    \item \textbf{Left-to-right inference}: sequentially unmasks tokens from left to right, consistent with the order in which residual k-means assigns SID tokens.
\end{itemize}

Table~\ref{tab:inference-strategies} reports the results of these experiments. We find that both greedy and left-to-right inference substantially outperform random inference, with greedy inference achieving slightly better performance than the fixed left-to-right order.

% As the training of the MaskGR doesn't have any inductive bias toward any particular order such as left-to-right in autoregressive training, it allows more flexibility in selecting the set of tokens to unmask during the inference \cite{zheng2024reparameterized, kim2025train}. In this section, we experiment with three choices for selecting the set of tokens. (1) Randomly selecting the set of tokens to unmask as given by vanilla MDM inference (2) Greedily selecting the tokens according to the uncertainty of values at the unmasking positions. The uncertainty at a masked SID position is determined using the difference between the probability of 1st and 2nd most likely assignment. (3) Selecting the tokens in left-to-right order. As the SIDs generated by residual K-means clustering have a natural left-to-right inductive bias, we also investigate the performance of fixed left-to-right order. 

% We report our results in Table~\ref{tab:inference-strategies}. We observe that greedy inference and left-to-right inference outperforms random inference significantly. Additionally, we observe that greedy inference performs slightly better than left-to-right inference. 

\begin{table}[ht]
    \centering
    % \small                          % choose \small, \footnotesize, or \scriptsize
    \setlength{\tabcolsep}{8pt}     % column spacing
    \renewcommand{\arraystretch}{1.15} % row spacing
    \caption{Performance of different inference strategies for MaskGR. By default, MaskGR uses the Greedy strategy.}
    \begin{tabular}{l|cccc}
        \toprule
        \textbf{Inference Method} & \textbf{R@5} & \textbf{R@10} & \textbf{N@5} & \textbf{N@10} \\
        \midrule
        MaskGR + Random         & 5.01 & 7.54 & 3.27 & 4.09 \\
        MaskGR + Left-to-right   & 5.31 & 8.09 & 3.46 & 4.37 \\
        MaskGR  (Greedy)         & 5.38 & 8.15 & 3.51 & 4.41 \\
        \bottomrule
    \end{tabular}
    \label{tab:inference-strategies}
\end{table}

\subsection{Q5. Extension via Dense Retrieval}

In this section, we present the experimental results of extending MaskGR with dense retrieval. The MaskGR with dense retrieval unifies the MaskGR's SID generation capabilities with dense retrieval as described in Section~\ref{sec:madrec-dense}.  We use the same 4096-dimensional Flan-T5-XXL  embeddings that we used for SID assignment as the items' text embeddings. We construct the predicted dense embedding of an item by combining the output of the 4th layer of our encoder model and projecting it to 4096 dimensions (matching the text embedding size) through a one-layer MLP with a hidden dimension of 256. Since both the text and predicted embeddings are high-dimensional, MLP with full-rank weights would introduce over one million parameters per layer. Therefore, we use an MLP with low-rank weights of rank $32$. During training, with a probability of $\beta = 0.2$, we mask all SIDs of an item to promote learning at the item-level abstraction. In the unified retrieval setup—where SID generation and dense retrieval are jointly leveraged—MaskGR first generates 20 beams using beam search and then re-ranks them based on their dense retrieval scores to obtain the top-10 candidates.

We report our results on the Beauty dataset in Table~\ref{tab:retrieval_comparison}. The results indicate that integrating MaskGR with dense retrieval and unified retrieval both improve performance compared to the generative retrieval baseline, while the two enhanced variants perform comparably to each other. Perhaps more importantly, these observations exemplify that MaskGR is general enough to be compatible with auxiliary methods developed for improving performance in AR modeling for GR. 
 
\begin{table}[h]
    \centering
    \setlength{\tabcolsep}{12pt}     % column spacing
    \caption{Performance comparison of MaskGR combined with different retrieval methods. By default, MaskGR uses  the generative retrieval strategy outlined in Section \ref{madrec:inference}.}
    \label{tab:retrieval_comparison}
    \begin{tabular}{l|cccc}
        \toprule
        \textbf{Method} & \textbf{R@5} & \textbf{R@10} & \textbf{N@5} & \textbf{N@10} \\
        \midrule
        MaskGR  & 5.38 & 8.15 & 3.51 & 4.41 \\ %+ Generative Retrieval
        MaskGR + Dense Retrieval & 5.41 & 8.50 & 3.53 & 4.45 \\
        MaskGR + Unified Retrieval & \textbf{5.43} & \textbf{8.59} & \textbf{3.54} & \textbf{4.47} \\
        \bottomrule
    \end{tabular}
\end{table}

% \subsection{Limitations}

% To understand the MaskGR's effectiveness in capturing semantic information from the semantic IDs extracted from the item embeddings, we perform two additional experiments on the Beauty dataset where we systematically remove the provided semantic information. In the first experiment, instead of using SIDs generated using item embeddings, we use random SIDs. In the second experiment, instead of modelling the probability distribution over the SID sequence like in MaskGR, we consider modelling the probability distribution over the item IDs directly. We compare the performance of MaskGR with random SIDs and item IDs in Table Table~\ref{}.

% \section{Abalation}

% \begin{itemize}
%     \item Different types of percentage of training data
%     \item Different inference strategies
%     \item Better coarse gained performance (graph)
%     \begin{itemize}
%         \item The performance gap improves as higher $k$.
%         \item SID-wise performance
%     \end{itemize}
%     \item Using temperature
%     \item Changing the inference steps
%     \item Different number of SIDs 
    
%     \item Different types of encoding (Absolute vs rotation)
%     % \item LIGER type dense embedding-based method
% \end{itemize}

\section{Related work}

\textbf{Generative Recommendation.} %Generative Recommendation (GR) has 
Advances in generative models in NLP and computer vision have inspired a number of innovations in modeling user interaction sequences in recent years. GRU4Rec \cite{jannach2017recurrent} and SASRec \cite{kang2018self} applied a gated recurrence unit and decoder-only transformer, respectively, to autoregressively predict the next interacted item, while BERT4Rec \cite{sun2019bert4rec} trained a BERT-style encoder using masked language modeling.
%BERT4Rec is most similar to our work, although it uses raw item IDs instead of SIDs and uses a fixed masking ratio.
More recently, a plethora of works have aimed to leverage the power of Large Language Models (LLMs) for sequential recommendation. Such approaches can be categorized as either {\em LLM-as-Recommender} or {\em LLM-as-Enhancer}. The former approaches treat the LLM itself as the recommendation system, which has achieved promising results \cite{li2023text, cao2024aligning, zheng2024adapting, bao2023tallrec} but faces  challenges in teaching the LLM the item corpus and collaborative signal. Conversely,  LLM-as-Enhancer treats the LLM as an auxiliary source of information to aid more traditional recommendation systems. This includes using LLMs to produce embeddings to replace item embedding tables \cite{sheng2024language, yuan2023go, hou2022towards}, distill semantic knowledge into smaller recommenders \cite{xu2024slmrec, liao2024llara, zhang2025recommendation, xi2024towards, gao2023chat}, and generate synthetic data for training \cite{liu2024recprompt}. However, arguably the most popular LLM-as-Enhancer paradigm entails using the LLM to produce embeddings for deriving SIDs.

% integrating text and SID tokens LC-Rec \cite{zheng2024adapting}

\noindent \paragraph{Generative Recommendation with SIDs.} 
% The GR with SIDs paradigm has garnered widespread attention by offering a performant means to incorporate both semantic and collaborative signals in item representations that require, in theory, exponentially fewer dense parameters than raw ID embedding tables.
% has emerged as one of the most popular approaches for leveraging these models \cite{rajput2023recommender}. 
Since its introduction in VQ-REC \cite{hou2023learning} and TIGER \cite{rajput2023recommender}, much of the work in GR with SIDs  has focused on improving item tokenization, via methods such as contrastive learning \cite{zhu2024cost}, incorporating collaborative signals  \cite{xiao2025progressive, hua2023index, qu2025tokenrec, wang2024learnable, wang2024eager, liu2024end}, directly using LLMs \cite{tan2024idgenrec, jin2023language} and  encouraging more uniform SID distributions \cite{kuai2024breaking}.
% , and tokenizing $n$-grams of SIDs \cite{houactionpiece}.
Other works have conditioned SID generation on user profiles \cite{paischer2024preference}, merged GR with dense retrieval \cite{yang2024unifying, yang2025sparse}, and employed tree-based decoding \cite{feng2022recommender, si2024generative}. 
However, all of these works adopt the  AR training strategy from \cite{rajput2023recommender}. 
% variety of works have innovated on the GR with SIDs framework \cite{}. However, the majority 
RPG \cite{hou2025generating} proposes to use product-quantized SIDs to enable decoding the SID tokens for a single item in parallel, however this approach requires decoding items sequentially and needs many SIDs per item to see performance gains.
% but still uses autoregressive training, and and requires many (64) SID token per item to achieve SOTA performance. 
In contrast, our method allows for parallel decoding with {\em any} SIDs, achieving SOTA even with the standard residual-quantized SIDs.
Outside of GR, \cite{ren2024non} and \cite{valluri2024scaling} employ non-AR, non-diffusion generative models for reranking and information retrieval, respectively.

% MaskGR can also be used with PQ SIDs, which may enhance parallel decoding performance in future work -- I don't think we should mention this bc reviewer will ask why we didn't try this
% \cite{yang2024unifying, }
% integrating text and SID tokens LC-Rec \cite{zheng2024adapting}

% LRMs...

% TIGER: \cite{rajput2023recommender}

% \textbf{GR with SIDs.} 

% Since its introduction by ..., GR with SIDs has drawn vast attention. ...
% However, all of these works default to autoregressive training. 
% ... uses ... Yet, ... 

% Thus far, the majority of research on GR with SIDs has focused on injecting additional information into model, especially by tokenizing items according to collaborative in addition to semantic signals \cite{zhu2024cost, xiao2025progressive, } and utilizing dense embeddings during next item decoding \cite{yang2024unifying, }. Other works have studied to how to encourage a more balanced distribution of SIDs and how to speed up inference  with the help of a novel tokenization strategy.

% SIDE \cite{ramasamy2025side} new quantization technique and using multiple content embeddings per item.
% bert4rec

% CaDiRec: \cite{Cui_2024}

% \textbf{Diffusion.}

\noindent \paragraph{Diffusion for Recommendation.}
% \url{https://github.com/tingruew/diffusionmodels-in-recsys} \\
% \url{https://arxiv.org/pdf/2410.13117}
The successes of diffusion models in CV and NLP have spurred increasing interest in leveraging diffusion models for RecSys in recent years. 
% Several works have explored diffusion modeling in the context of collaborative filtering \cite{}, opting to diffuse over , which is 
% 
% Existing works can be grouped by the space that they diffuse over, i.e. the entity to which they add noise and learn to de-noise. 
Several works diffuse over the space of user interaction vectors \cite{wang2023diffusion, ma2024plug, walker2022recommendation, yu2023ld4mrec, hou2024collaborative, zhao2024denoising, walker2022recommendation} or the user-item interaction graph \cite{zhu2024graph, chen2024g, choi2023blurring}, but the dimension of these spaces grows with the item cardinality, raising scalability concerns.
Numerous other works apply continuous diffusion over the space of item embeddings, in both the traditional collaborative filtering \cite{li2025diffgraph} and sequential recommendation \cite{yang2023generate, li2023diffurec, li2025dimerec, wang2024conditional, yuan2025hyperbolic, mao2025distinguished, chen2025unlocking} contexts. 
% Among works dealing with sequential recommendation,  DreamRec \cite{yang2023generate} applies continuous diffusion conditioned on the user's sequence to denoise the target item embedding. ...
% DiffRec, DreamRec, DiffuRec, DimeRec, ADRec
% https://arxiv.org/pdf/2505.19544
% However, the majority of these works do not leverage semantic information, operating instead on purely collaborative embeddings.
\citet{cui2024cadirec} and \citet{hu2024generate} utilize semantic information by executing continuous diffusion over pretrained semantic embeddings.
% , however these approaches again rely on continuous diffusion. 
% Other works have applied continuous diffusion models for out-of-corpus item recommendation \cite{}, cross-domain recommendation \cite{},
% mention diffusion sid paper here
DDSR \cite{xie2024breaking} uses a discrete diffusion process over SIDs to predict next items, but employs this in an AR manner.

\section{Conclusion}
We present a novel modeling paradigm for GR with SIDs inspired by recent advancements in masked diffusion modeling in NLP. Empirically, our framework brings several advantages over standard AR modeling, including improved generalization in data-sparse settings, inference efficiency, and coarse-grained recall. Our framework is also simple and general enough to incorporate innovations from AR SID modeling, such as incorporating dense retrieval. Exploring novel design choices to combine other such auxiliary techniques with MaskGR presents an interesting direction for future research. We also envision that user sequence modeling with masked diffusion can be further improved by a more sophisticated training and inference guidance strategy (e.g., classifier-free/classifier-based guidance \cite{schiff2024simple} or error correction via remasking \cite{wang2025remasking, von2025generalized}).

\bibliography{refs}

%%
%% If your work has an appendix, this is the place to put it.
\appendix

\section{Experiments}

\subsection{Additional Experimental Details}\label{app:experiments}

\noindent \textbf{Dataset statistics.} We summarize the detailed statistics for our datasets (Amazon Beauty, Sports, Toys, and MovieLens-1M) in Table~\ref{tab:data}.

\begin{table}[h]
    \centering
    \caption{Dataset statistics after preprocessing.}
    \label{tab:data}
    \setlength\tabcolsep{12pt}
    % \renewcommand{\arraystretch}{1.1}
    % \scalebox{0.875}{
    \begin{tabular}{l|rrrr}
        \toprule
        \textbf{Dataset} & \textbf{Beauty} & \textbf{Toys} & \textbf{Sports} & \textbf{ML-1M} \\
        \midrule
        \# Users & 22,363 & 19,412 & 35,598 & 6,040  \\
        \# Items & 12,101 & 11,924 & 18,357 & 3,416 \\
        \# Interactions & 198,502 & 167,597 & 296,337 & 999,611 \\
        \# Avg. Length & 8.88 & 8.63 & 8.32 & 165.50 \\
        Sparsity & 99.93\% & 99.93\% & 99.95\% & 95.16\% \\
        \bottomrule
    \end{tabular}%}
\end{table}

\noindent \textbf{Implementation details.}
Our implementation of MaskGR is based on the GRID codebase presented in \cite{grid}. We use the AdamW optimizer with a learning rate of 0.005, a weight decay of 0.001, a batch size of 8192, and early stop all experiments using validation recall@10. We use the implementation of TIGER \cite{rajput2023recommender} provided in GRID. 
We use the implementation of LIGER \cite{yang2024unifying} using the public repository released by the paper's authors.
Both TIGER and LIGER  use 4 SIDs per item, including the deduplication token, consistent with their original implementations and TIGER's optimal setting for the Amazon datasets \cite{grid}. All SIDs are assigned by executing residual $k$-means clustering with $c=256$ clusters per layer on mean-pooled Flan-T5-XXL embeddings. For baselines, we use SASRec results reported by Rajput et al. \cite{rajput2023recommender}, and BERT4Rec, DreamRec, and CaDiRec results from Cui et al. \cite{cui2024cadirec}.

% {\color{red} rope embeddings? temperature? early stopping/num training epochs? sasrec bert4rec results taken from tiger paper? dreamrec and cadirec implementations?}

We execute all experiments on nodes with four 16-GB NVIDIA V100 GPUs or four 40 GB NVIDIA A100 GPUs.

\subsection{Additional Experimental Results}

In Table \ref{tab:recall10-table} we share the NDCG@10 and Recall@10 metrics for the same experiments as in Table \ref{tab:main-result-horizontal-clean}.

\begin{table*}[htb]
    \centering
    \caption{Comparison of the performance of MaskGR with other GR methods on multiple datasets.}
    \label{tab:recall10-table}
    \begin{tabular}{l *{8}{c}} 
        \toprule
        \multirow{2}{*}{\textbf{Method}} & \multicolumn{2}{c}{\textbf{Beauty}} & \multicolumn{2}{c}{\textbf{Sports}} & \multicolumn{2}{c}{\textbf{Toys}} & \multicolumn{2}{c}{\textbf{ML-1m}} \\
        \cmidrule(lr){2-3} \cmidrule(lr){4-5} \cmidrule(lr){6-7} \cmidrule(lr){8-9}
        & \textbf{R@10} & \textbf{N@10} & \textbf{R@10} & \textbf{N@10} & \textbf{R@10} & \textbf{N@10} & \textbf{R@10} & \textbf{N@10} \\
        \midrule
        SASRec & 6.05 & 3.18 & 3.50 & 1.92 & 7.12 & 4.32 & 16.89 & 7.72 \\
        BERT4Rec & 6.01 & 3.08 & 3.59 & 1.81 & 6.65 & 3.68 & 20.56 & 11.12 \\
        DreamRec & 6.87 & 3.52 & 3.74 & 1.91 & 6.43 & 4.02 & 20.29 & 10.47 \\
        CaDiRec & \underline{7.18} & \underline{3.86} & \underline{4.25} & \underline{2.33} & \underline{7.85} & \underline{4.41} & \underline{22.82} & \underline{12.51} \\
        TIGER & 6.33 & 3.54 & 3.61 & 2.03 & 6.63 & 3.61 & 19.97 & 10.13 \\
        LIGER & \underline{7.52} & \underline{4.14} & \underline{4.27} & 2.30 & 6.25 & 3.52 & 20.58 & 10.81 \\
        \midrule 
        \textbf{MaskGR} & \textbf{8.15} & \textbf{4.41} & \textbf{4.54} & \textbf{2.49} & \textbf{8.46} & \textbf{4.45} & \textbf{23.96} & \textbf{13.45} \\
        (+ {Improv. (\%)}) & +8.7 \% & +6.5 \% & \textbf +6.3 \% & +6.9 \% & +7.8 \% & +0.9 \% & +5.0 \% & +7.5 \% \\
        \bottomrule
    \end{tabular}
\end{table*}

% \received{7 October 2025}
% \received[revised]{12 March 2009}
% \received[accepted]{5 June 2009}

\end{document}